\documentclass[lettersize,journal]{IEEEtran}
\usepackage{amsmath,amsfonts}
\usepackage{algorithmic}
\usepackage{algorithm}
\usepackage{array}
\usepackage[caption=false,font=normalsize,labelfont=sf,textfont=sf]{subfig}
\usepackage{textcomp}
\usepackage{stfloats}
\usepackage{url}
\usepackage{verbatim}
\usepackage{graphicx}
\usepackage{cite}
\usepackage{tabularx}
\usepackage{multirow}
\usepackage{float}
\usepackage{booktabs,ragged2e}
\usepackage{hyperref}
\usepackage{afterpage}
\usepackage{pdfpages}
\divide\pdfpxdimen by 300

\usepackage{lineno} 

\begin{document}

\title{Location-aware Adaptive Normalization: A Deep Learning Approach For Wildfire Danger Forecasting}

\author{Mohamad Hakam Shams Eddin, Ribana Roscher,~\IEEEmembership{Member,~IEEE} and Juergen Gall,~\IEEEmembership{Member,~IEEE}

\thanks{Manuscript created December, 2022; revised April 2023; This work was funded by the Deutsche Forschungsgemeinschaft (DFG, German Research Foundation) – SFB 1502/1–2022 - Projectnumber: 450058266.}
\thanks{Corresponding author: Mohamad Hakam Shams Eddin. The authors are with the University of Bonn, Germany (e-mail: \{shams, gall\}@iai.uni-bonn.de), the Lamarr Institute for Machine Learning and Artificial Intelligence, and the Research Center Jülich (e-mail: ribana.roscher@uni-bonn.de).}
}



\maketitle


\begin{abstract}
Climate change is expected to intensify and increase extreme events in the weather cycle. Since this has a significant impact on various sectors of our life, recent works are concerned with identifying and predicting such extreme events from Earth observations. With respect to wildfire danger forecasting, previous deep learning approaches duplicate static variables along the time dimension and neglect the intrinsic differences between static and dynamic variables. Furthermore, most existing multi-branch architectures lose the interconnections between the branches during the feature learning stage. To address these issues, this paper proposes a 2D/3D two-branch convolutional neural network (CNN) with a Location-aware Adaptive Normalization layer (LOAN). Using LOAN as a building block, we can modulate the dynamic features conditional on their geographical locations. Thus, our approach considers feature properties as a unified yet compound 2D/3D model. Besides, we propose using the sinusoidal-based encoding of the day of the year to provide the model with explicit temporal information about the target day within the year. Our experimental results show a better performance of our approach than other baselines on the challenging FireCube dataset.
The results show that location-aware adaptive feature normalization is a promising technique to learn the relation between dynamic variables and their geographic locations, which is highly relevant for areas where remote sensing data builds the basis for analysis. The source code is available at \href{https://github.com/HakamShams/LOAN}{\color{violet}https://github.com/HakamShams/LOAN}.
\end{abstract}


\section{Introduction}\label{sec:I}
\IEEEpubidadjcol
\IEEEPARstart{T}{here} is a general expectation that weather and climate extremes will change their patterns and frequencies in the future \cite{Ren, lausier2018overlooked, perkins2020increasing, samuels2018evaluation}. This is particularly the case for the Mediterranean region, which has been identified as a hot spot for climatic changes \cite{zittis2019multi, esd-11-161-2020, esd-13-749-2022}. Because extreme weather events can impose short- and long-term risks in our Earth system, predicting these risks such as droughts, windstorms, and wildfires has become recently more relevant. In particular, wildfire forecasting constitutes one of the open challenges for risk assessment and emergency response \cite{thompson2019risk, Coogan2019ScientistsWO, Moreira_2020}. Wildfire forecasting refers to the task of fire-susceptibility mapping using key remote sensing, meteorological, and anthropogenic variables \cite{jain2020review}. Building an integrated modeling system of the Earth should also consider wildfire events to comprehend the origin of past patterns better and predict future ones \cite{rs9111131}. Unlike typical prediction tasks, understanding when weather conditions have a high tendency to cause fire events addresses more complexities; among these are the stochastic nature of fire events \cite{Prapas2} and fire drivers, which are time-dependent and inter-correlated across variables \cite{hantson2016status}. Moreover, the prediction model should consider difficulties like a high false positive error rate, uncertainty, and class imbalance. 

In recent years, many works leveraged classical machine learning approaches to solve the task \cite{jain2020review}. More recently, deep learning methods \cite{reichstein2019deep} have become popular since they can handle large multivariate datasets more efficiently and are able to learn highly complex relationships between observations and the predicted outcome. 
In the context of wildfire danger forecasting, Prapas et al.~\cite{Prapas1} and Kondylatos et al.~\cite{Prapas2} proposed to use recurrent neural networks in combination with 2D convolutions to exploit both temporal and spatial context. These approaches, however, do not distinguish between the different input variables. Static variables like elevation, which barely change over time, are simply copied and concatenated with dynamic variables like surface temperature. This results not only in a highly redundant input to the network, but it also neglects strong causal effects between static and dynamic variables. For instance, the surface temperature strongly depends on the geographical location, which is described by static variables.  

In this work, we thus propose a convolutional neural network for wildfire danger forecasting that handles static and dynamic variables differently. Since the static variables do not change over time, they are processed by a branch consisting of 2D convolutions while the dynamic variables are processed by the second branch with 3D convolutions as illustrated in Fig.~\ref{fig:1}. To address the causal effect of static variables on dynamic variables, we   introduce \textit{feature modulation} for the dynamic variables where the modulation parameters are generated dynamically and conditionally on the geographical location. We thus name this method Location-aware Adaptive Normalization (LOAN). In addition, we encode the date of the forecasting during a year by an absolute time encoding based on the sinusoidal encoding~\cite{vaswani2017attention}. Both LOAN and the time encoding can be implemented as plugin layers in different deep learning architectures. We view our model as a generic architecture that can be used for other time-dependent forecasting tasks with static and dynamic variables.
\IEEEpubidadjcol
We conduct extensive experiments on the FireCube dataset~\cite{prapas_ioannis_2022_6475592} where our approach outperforms previous works. We achieve an overall improvement of up to {\boldmath $5.72\%$} in precision, {\boldmath $3.24\%$} in F1-score, {\boldmath $0.63\%$} in AUROC, and {\boldmath $1.15\%$} in OA on the test set. 

The rest of this paper is organized as follows. Section \ref{sec:II} reviews the related literature. Section \ref{sec:III} provides information about the used dataset. The proposed method is described in detail in Section \ref{sec:IV}. The experimental results and ablation study are provided in Section \ref{sec:V} and Section \ref{sec:VI}, respectively. Finally, conclusions and outlook are given in Section \ref{sec:VII}.
\section{Related Works}\label{sec:II}
\subsection{Wildfire Danger Forecasting}\label{sec:II-A}
Wildfire forecasting or wildfire-susceptibility mapping from remote sensing and Earth observations data is a very important topic for wildfire management\cite{jain2020review}. We briefly review some prior related works in this direction. Iban et al.~\cite{IBAN2022101647}, Pham et al.~\cite{sym12061022}, and Gholami et al.~\cite{Gholami2020WhereTS} relied on traditional machine learning approaches to generate susceptibility maps. Shang et al.~\cite{Shang2020SpatiallyExplicitPO} and Mitsopoulos and Mallinis~\cite{mitsopoulos2017data} utilized Random Forests classifiers (RF). In their works, they studied the importance of biotic and abiotic predictors for wildfire forecasting. Jiang et al.~\cite{9672044} proposed a deep learning approach based on a Multi-Layer Perceptron (MLP) and included a comparison with traditional machine learning algorithms. In Le et al.~\cite{LE2021101300}, a similar MLP-based approach was presented to generate a forest fire danger map. Zhang et al.~\cite{zhang2019} used a convolutional neural network (CNN) and extended their work later to predict fire susceptibility at the global level \cite{zhang2021}. Other works with CNN were conducted in Bj{\aa}nes et al.~\cite{BJANES2021101397} and Bergado et al.~\cite{bergado2021predicting}. Furthermore, Huot et al.~\cite{9840400} approached the problem as a scene classification task using U-Net models to predict wildfire spreading. Their approach operates directly on the whole scene. A similar approach based on a U-Net++ model for global wildfire forecasting was proposed in Prapas et al.~\cite{prapas2022deep}. Yoon and Voulgaris \cite{Yoon2022MultitimePO} presented an approach that relies on a recurrent network with Gated Recurrent Units (GRU) to model past observations and on a CNN to predict wildfire probability maps for multiple time steps. More recently, Prapas et al.~\cite{Prapas1} and Kondylatos et al.~\cite{Prapas2} proposed to use LSTM-based (Long-Short Term Memory) approaches. They exploited both temporal and spatio-temporal context by applying recurrent LSTM and ConvLSTM models. They did not consider the whole scene at once but rather the classification of one pixel at a time (pixel-level).

Unlike these works, we do not treat all observation variables in the same way, but we propose a deep learning model that handles different types of variables in separated 2D and 3D CNN branches. In contrast to a ConvLSTM, which models spatial and temporal relations separately, a 3D CNN models spatio-temporal relations jointly. We assume that the dataset contains static and dynamic variables, which we argue is the case for most datasets. Similar to Prapas et al.~\cite{Prapas1} and Kondylatos et al.~\cite{Prapas2}, we also formulate the problem as pixel-level classification taking into account the spatio-temporal local context around the target pixel.

\subsection{Multi-Branch Neural Networks}\label{sec:II-B}
When deep learning is applied to potentially multi-source remote sensing-based Earth observation data, multi-branch neural networks are a commonly used framework. This is mainly because such networks enrich representation learning and provide discriminative learning perspectives of the input variables \cite{9606819}. In addition, an important aspect of the multi-branch design is the capability to adapt some parts of the model to a specific type of input. The general framework generates features from each branch and fuses these features in the network to obtain a unified feature vector. This fused representation is used as input to the subsequent layers. In Gaetano et al.~\cite{rs10111746}, a two-branch 2D CNN network was proposed to handle panchromatic information along with a multi-spectral one for image classification. Tan et al.~\cite{TAN2020358} reduced the depth of a semantic segmentation classifier by applying consecutive blocks, each containing three CNN branches. A similar objective can be found in Zhao et al.~\cite{9606819}, where the network complexity was reduced via weight sharing and self-distillation (SD) embedding. In this way, only the main branch is used during inference, which inherits the knowledge of trained subbranches and has a close performance to an ensemble model. For hyperspectral image classification, Xu et al.~\cite{8356713} introduced a model called Spectral–Spatial Unified Network (SSUN). In their model, spectral features are learned by a grouping-based LSTM, and spatial features are learned by a 2D CNN. Shen et al.~\cite{8835034} used separated spectral and spatial convolutional branches for hyperspectral input ($\text{S}^2$CDELM). They based their framework on the extreme learning machine (ELM). Unlike common backpropagation algorithms, they used a single hidden layer feed-forward model. A multi-branch architecture was also explored for image fusion. Liu et al.~\cite{8693668} proposed a two-stream CNN called (StfNet). They investigated the task of spatio-temporal image fusion. Their network takes a coarse image input along with its neighboring images to predict the reconstructed fine image. Some works adapted a multi-branch architecture to construct a multi-scale feature vector. In Gan et al.~\cite{8938811}, a dual-branch CNN with different filter kernel sizes was used as an autoencoder. Thus, the input image could be processed at different scales. Tang et al.~\cite{9767822} proposed a multi-scale Gaussian pyramid to handle hyperspectral input. They used the Gaussian pyramid to obtain multi-scale images which are then processed by ResNet modules \cite{8127330}. In this way, spatial features can be learned at different scales. They further used a second branch, which performs a discrete wavelet transform on the spectral input followed by an LSTM module \cite{LSTM}. The spatial and spectral features are then fused and processed by an MLP to obtain the final classification result. For short-term multi-temporal image classification, Zheng et al.~\cite{9573468} addressed the task through a Multi-temporal Deep Fusion Network (MDFN). In their framework, the LSTM-based branch is used to learn temporal-spectral features. At the same time, temporal-spatial and spectral-spatial information is learned via a joint 3D-2D CNN with two branches. Furthermore, some works employed attention mechanisms with multi-branch architectures \cite{rs12030582,9781404,9833261}. When attention is applied, it drives the model to focus more on regions of interest. The former described methods \cite{rs10111746, TAN2020358, 8356713, 8835034, 8693668, 8938811, 9767822, 9573468, rs12030582} except of Zhao et al.~\cite{9606819}, Zhu et al.~\cite{9781404}, and Deng et al.~\cite{9833261} did not consider linking information between branches during the feature learning stage. This limits the gradient flow and disentangles the correlations between the learned features. 

This paper proposes an architecture composed of two CNN branches for a forecasting task. A 2D branch is used to learn spatial features from static variables. At the same time, a 3D branch is used to learn spatio-temporal features from dynamic variables, which vary along the input temporal dimension. The branches are further linked via adaptive modulation layers to model the causal effects of static variables on dynamic variables.
\begin{figure*}[ht!]
    \centering
    \includegraphics[width=0.98\textwidth]{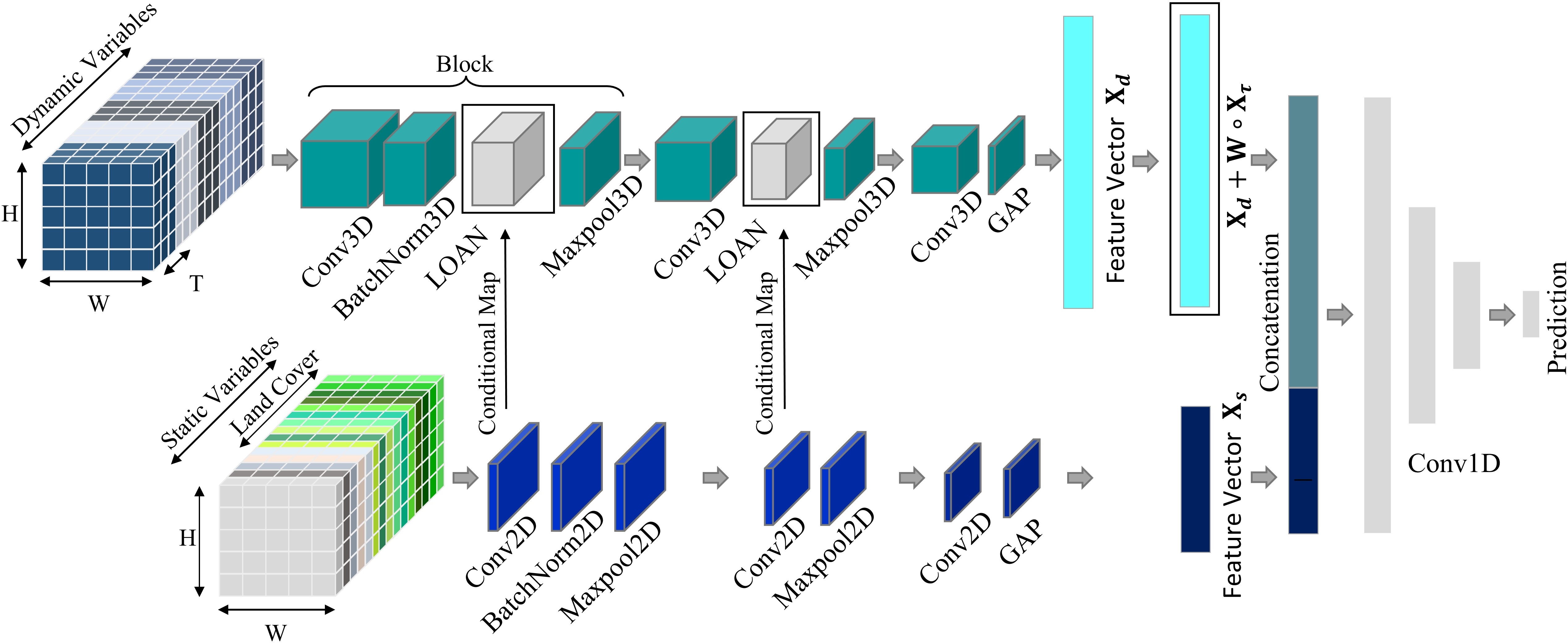}
    \caption{Overview of the proposed approach. Our network handles static and dynamic variables by two branches. For the static branch, we use 2D convolutions whereas we use 3D convolutions for the dynamic branch since the dynamic variables change over time. Since dynamic variables like surface temperature strongly depend on static variables like elevation, we modulate the dynamic features conditioned on the static features at two blocks (LOAN). This makes the model location-aware since the static variables contain geographic data like land cover. The feature vector $\textbf{X}_{d}$ from the dynamic branch is combined with a weighted temporal encoding vector $\textbf{X}_{\tau}$ before it is concatenated with the static feature vector $\textbf{X}_{s}$. The concatenated vector serves as input to fully connected layers that predict the probability of a wildfire.}
    \label{fig:1}
\end{figure*}

\subsection{Conditional Normalization}\label{sec:II-C}
Since the introduction of normalization techniques in deep learning, they became a basic building block in many state-of-the-art models. Common normalization methods include batch normalization \cite{ioffe2015batch}, group normalization \cite{wu2018group}, instance normalization \cite{Ulyanov2016InstanceNT}, and layer normalization \cite{Ba2016LayerN}. It has been shown empirically that normalization layers help with model optimization and regularization. Through normalization layers, the activation maps inside the model are normalized to follow a normal distribution with zero mean.
After that, the normalized activation maps are modulated or denormalized by learnable affine transformation parameters. 
These parameters vary across channels and are learned based on the running training statistics together with the model parameters. Therefore, such a normalization method is called unconditional. Compared to popular unconditional normalizations, there exist conditional normalization techniques which aim to learn affine parameters conditionally on external input. In the field of computer vision, conditional normalization is often used for image synthesis and style transformation \cite{8953676, 8578168, huang2017arbitrary, 9578482, Noord_2017_ICCV, 9156510, 9419767, 9578869, lyu2021detailed, 9707002, Chen2019OnSM, Huang2017ArbitraryST, OASIS, DPOASIS}. More recently, Mar{\'\i}n and Escalera \cite{rs13193984} adapted the conditional normalization from Wang et al.~\cite{8578168} and Park et al.~\cite{8953676} to generate high-resolution satellite images. We use the conditional normalization in a very different way than \cite{rs13193984,8578168,8953676}. While these works focus on synthesizing an image using the segmentation map as conditional input, we aim to modulate dynamic features conditioned on static features. 

\subsection{Temporal Positional Encoding}\label{sec:II-D}
A plethora of studies exist about temporal modeling in remote sensing \cite{shi2015convolutional, ijgi7040129, Pelletier2019Temporal, rs13234822}.
Recently, the self-attention model, also known as Transformer and first presented by Vaswani et al.~\cite{vaswani2017attention} for natural language processing, has become a natural choice to handle sequential data, which includes a positional encoding. In the field of remote sensing, many works showed the benefits of adapting positional encoding for time-dependent image classification (Garnot et al.~\cite{garnot2020satellite}, Garnot and Landrieu \cite{10.1007/978-3-030-65742-0_12}), panoptic segmentation (Garnot and Landrieu \cite{9711189}), and image generation (Dress et al.~\cite{drees2022time}). In their work, each image was given an encoded time vector according to its position with respect to a reference point, i.e., the first acquisition time step. Nyborg et al.~\cite{NYBORG2022301} used the calendar time (day of the year) to provide positional information within the sequence. They also proposed to learn or estimate time shifts between geographically distant regions to enhance the generalization further. In another work, Nyborg et al.~\cite{9857456} used the thermal time, which is obtained by accumulating daily average temperatures over the growing season, for crop classification. In general, transformer-based approaches require the positional encoding since the temporal information is otherwise lost within a transformer model. While 3D CNNs consider the temporal order of the input such that a positional encoding is not necessary, we show in this work that adding an absolute temporal encoding is also useful for time-dependent forecasting.    
\section{Dataset}\label{sec:III}
There are only very few publically available datasets for wildfire forecasting and they differ significantly in the observational variables, the spatial and temporal resolutions, and the task that needs to be addressed. A related dataset is the Next Day Wildfire Spread dataset \cite{9840400} where the task is to predict wildfire spread. It is formulated as a scene classification task and not as a pixel-wise wildfire forecasting task as it is proposed in the FireCube dataset \cite{prapas_ioannis_2022_6475592} and addressed in this work. The FireCube dataset was first published in Prapas et al.~\cite{Prapas1} and extended later in Kondylatos et al.~\cite{Prapas2}. It includes multivariate spatio-temporal data streams with 90 variables from the years $2009$-$2021$ with a resolution of $1\text{ km} \times 1\text{ km} \times 1\text{ day}$. The area is $1253\text{ km} \times 983\text{ km}$, covering parts of the Eastern Mediterranean. The observational variables include meteorological data \cite{essd-13-4349-2021}, satellite-derived products \cite{didan, wan2015mod11a2}, topographic features \cite{bashfield2011continent}, human-related activities \cite{tatem2017worldpop}, and historical fire records \cite{SANMIGUELAYANZ201319, GIGLIO201631}. In addition, Copernicus Corine Land Cover (CLC) \cite{Buettner2014} and Fire Weather Index (FWI) \cite{Steinfeld} are provided.
The target is to predict for each pixel if a wildfire will ignite and become large ($>0.3~{\text{km}}^2$) in the next day. The task is equivalent to binary classification, where the positive class represents a wildfire event.

Since wildfire forecasting is essentially an imbalanced classification task, the authors of \cite{Prapas2} extracted samples as follows:
For a target day $T{+}1$, the static variables form a patch of $25\text{ km} \times 25\text{ km}$ centered around the target pixel at day $T$. In contrast, the dynamic variables consist of $25\text{ km } \times 25\text{ km } \times 10\text{ days}$ time series of observations from days $T{-}9$ until $T$. For each positive sample, a few negative samples from different locations are sampled. Although the negatives are from different locations, they are sampled from regions that have a similar land cover distribution as the positive samples to make the task more difficult. 

Overall the dataset includes $71471$ samples for training ($13518$ positives and $57953$ negatives for the years $2009$-$2018$), $6430$ samples for validation ($1300$ positives and $5130$ negatives for the year $2019$) and $42820$ samples for testing ($1228$ positives and $4860$ negatives for the year $2020$ and $4407$ positives and $32325$ negatives for the year $2021$). The year $2021$ in the test set contains an extreme wildfire season in Greece \cite{atmos13030475, Prapas2}. The extracted samples are available in \cite{prapas_ioannis_2022_6528394}.

In this paper, we use from the described dataset the same variables as in Kondylatos et al.~\cite{Prapas2}. This includes the following:
\begin{itemize}
\item{$15$ static variables:}
\begin{itemize}
    \item{Digital elevation model (DEM) and Slope \cite{bashfield2011continent}.}
    \item{Distance to roads, distance to waterway, and population density \cite{tatem2017worldpop}.}
    \item{Copernicus Corine Land Cover variables representing the fractions of classes for each pixel. This gives $10$ variables per pixel \cite{Buettner2014}.}
\end{itemize}
\item{$10$ dynamic variables:}
\begin{itemize}
    \item{Day and night land surface temperature \cite{wan2015mod11a2}.}
    \item{Normalized difference vegetation index (NDVI) \cite{didan}.}
    \item{Soil moisture index \cite{cammalleri2017comparing}.}
    \item{Maximum $2$m temperature, maximum wind speed, minimum relative humidity, total precipitation, maximum $2$m dew point temperature, and maximum surface pressure \cite{essd-13-4349-2021}.}
\end{itemize}
\end{itemize}

\section{Methodology}\label{sec:IV}
\emph{Problem formulation}. Given a multivariate spatio-temporal data cube $\mathcal{C}(\{V, T, W, H\})$, where \textit{H} and \textit{W} are the spatial extensions of the cube, \textit{T} is the temporal extension in the past for the time series $1,2,\ldots,T$, and \textit{V} is the number of variables (static and dynamic), our aim is to learn a mapping function \textit{f} approximated by a neural network that can predict the probability $Y_{T{+}1}\in[0, 1]$ of a wildfire event to start at the center of $W \times H$ for the target day $T{+}1$:
\begin{equation}
\label{eq:1}
f : \mathcal{C}(\{V, T, W, H\}) \rightarrow\ Y_{T{+}1}\,.
\end{equation}
To achieve this, we propose a spatio-temporal 2D/3D CNN with two branches as illustrated in Fig.~\ref{fig:1}. First, the network design is introduced in Section~\ref{sec:IV-A}. Then, the Location-aware Adaptive Normalization layer (LOAN), which is the core of our work is explained in detail in Section~\ref{sec:IV-B}. Finally, Section~\ref{sec:IV-C} describes the integration of the absolute temporal encoding (TE) into the model. 

\subsection{2D/3D Two-Branch CNN}\label{sec:IV-A}
As shown in Fig.~\ref{fig:1}, our network consists of two branches that process dynamic and static variables, respectively. We denote the data cube with dynamic variables by $\mathcal{C}(\{V_d, T, W, H\})$
 and the data cube with static variables by $\mathcal{C}(\{V_s, W, H\})$. As in previous works, we normalize the input channel-wise to the range $[0,1]$.
 Since the static variables do not have a time component, we use 2D convolutions for the static branch and 3D convolutions for the dynamic branch. More in detail:      

\emph{Dynamic branch}. The dynamic branch takes the variables $V_d$ which vary over time as input. It consists of $3$ blocks; each block has a 3D convolution with a $3 \times 3$ kernel size followed by a ReLU activation function and a 3D max pooling layer. To reduce overfitting, we use global average pooling (GAP) \cite{lin2013network} at the end of the last block. We denote the feature vector learned from this branch as $\textbf{X}_d \in \mathbb{R}^{256}$.

\emph{Static branch}. In parallel to the dynamic branch, the static branch has a similar architecture. However, 2D convolutions are used instead of 3D ones. We denote the feature vector learned from this branch as $\textbf{X}_s \in \mathbb{R}^{128}$. Note that the dimensionality of the static feature vector is lower than the dimensionality of the dynamic feature vector since the input data cube is smaller.         

In a nutshell, the dynamic- and static branch are two functions $f_{d}$ and $f_{s}$, respectively:
\begin{align}
    f_{d} : \mathcal{C}(\{V_d, T, W, H\}) \rightarrow\ \textbf{X}_{d} \label{eq:2}\,, \\
    f_{s} : \mathcal{C}(\{V_s, W, H\}) \rightarrow\ \textbf{X}_{s}\,. \label{eq:3}
\end{align}

For the dynamic feature vector we add an absolute temporal encoding $\textbf{X}_\tau$, which will be described in Section \ref{sec:IV-C}. The two feature vectors are then concatenated and fed into 4 classification layers with 1D convolutions of kernel size 1. The layers reduce the dimensionality from 384 to 256, 128, 32, and 2. To reduce overfitting, we use dropout with a dropout probability $p=0.5$ for the 1D convolutional layers except the last two layers. Finally, a softmax activation is used after the last classification layer to predict the probability of a wildfire. In addition, we use a batch normalization layer \cite{ioffe2015batch} for the $1^{st}$ block of each branch. More implementation details are given in Section \ref{sec:V}.

For training, we use the binary cross entropy as loss function:
\begin{equation}
\label{eq:13}
-\frac{1}{N} \sum_{n=1}^{N}\left[\hat{Y}_{T{+}1}^{(n)}\log(Y_{T{+}1}^{(n)}) {+} (1{-}\hat{Y}_{T{+}1}^{(n)}) \log(1{-}Y_{T{+}1}^{(n)})\right],
\end{equation}
where $N$ denotes the batch size and $\hat{Y}_{T{+}1}^{(n)}\in\{0, 1\}$ is the true label for sample $n$.

In the following, we discuss the Location-Aware Adaptive Normalization (LOAN) that modulates the dynamic features based on the static features and the already mentioned absolute temporal encoding.    

\subsection{Location-Aware Adaptive Normalization (LOAN)}\label{sec:IV-B}
In general, dynamic variables are correlated with the geographic location, i.e., temperature and pressure change with elevation, soil moisture and NDVI vary with land cover, and humidity is correlated with some static variables like the waterway distance. Since the dynamic variables depend on the static variables and not vice versa, we aim to exploit this knowledge in our approach. This is done by learning to normalize the dynamic features based on the location-dependent static features. To this end, we propose a conditional normalization technique for remote sensing data called Location-aware Adaptive Normalization (LOAN).

We first describe a batch-normalization \cite{ioffe2015batch} where the activation map is normalized before it is modulated by scale $\gamma$ and bias $\beta$. Let ${z_d}_{}^{i} \in\ \mathbb{R}^{N \times K^{i} \times D^{i} \times W^{i} \times H^{i}}$ be an activation map 
in the $i\text{-th}$ block of the dynamic branch and ${z_s}_{}^{i} \in \mathbb{R}^{N \times K^{i} \times W^{i} \times H^{i}}$ be an activation map of the corresponding $i\text{-th}$ block in the static branch, where $K^{i}$ denotes the number of channels and $D^{i}$, $W^{i}$ and $H^{i}$ denote the depth, width, and height of the activation map $z^{i}$, respectively. 
Using the indices $n\in \{1, \ldots, N\}$, $k\in \{1, \ldots, K^i\}$, $t\in \{1, \ldots, D^i\}$, $w\in \{1, \ldots, W^i\}$, and $h\in \{1, \ldots, H^i\}$, the normalization of the dynamic branch is performed by the following equation:
\begin{equation}
\label{eq:4}
\left( \frac{{z_d}_{(n,k,t,w,h)}^i-\mu_{k}}{\sigma_{k}}\right)~.~\gamma_{k}+\beta_{k}\,,
\end{equation}
where ${z_d}_{(n,k,t,w,h)}^i$ is the activation before the normalization, $\mu_{k}$ and $\sigma_{k}$ are the computed mean and standard deviation of channel $k$, i.e., computed over the tensor $D^i\times W^i\times H^i$ and all samples $n$ in the batch, and $\gamma_{k}$ and $\sigma_{k}$ are the learnable modulation parameters. 

In our case, we aim to learn a modulation of the dynamic features ${z_d}_{(n,k,t,w,h)}^i$ at the $i\text{-th}$ block where the modulation parameters ${\gamma_s}_{(n,k,w,h)}^i$ and ${\beta_s}_{(n,k,w,h)}^i$ depend on the corresponding static features ${z_s}_{(n,k,w,h)}^i$:
\begin{equation}
\label{eq:8}
{z_d}_{(n,k,t,w,h)}^i~.~{\gamma_s}_{(n,k,w,h)}^i+{\beta_s}_{(n,k,w,h)}^i\,.
\end{equation}
In contrast to \eqref{eq:4}, the modulation parameters ${\gamma_s}_{(n,k,w,h)}^i$ and ${\beta_s}_{(n,k,w,h)}^i$ vary with respect to sample $n$ in the batch, the location $(w,h)$, and across channel $k$, but they are constant over time $t$. Furthermore, they are conditioned on the static features ${z_s}_{}^{i}$. We thus call ${z_s}_{}^{i}$ the conditional map for the modulation.

\begin{figure}[t]
    \centering
    \includegraphics[width=3.5in]{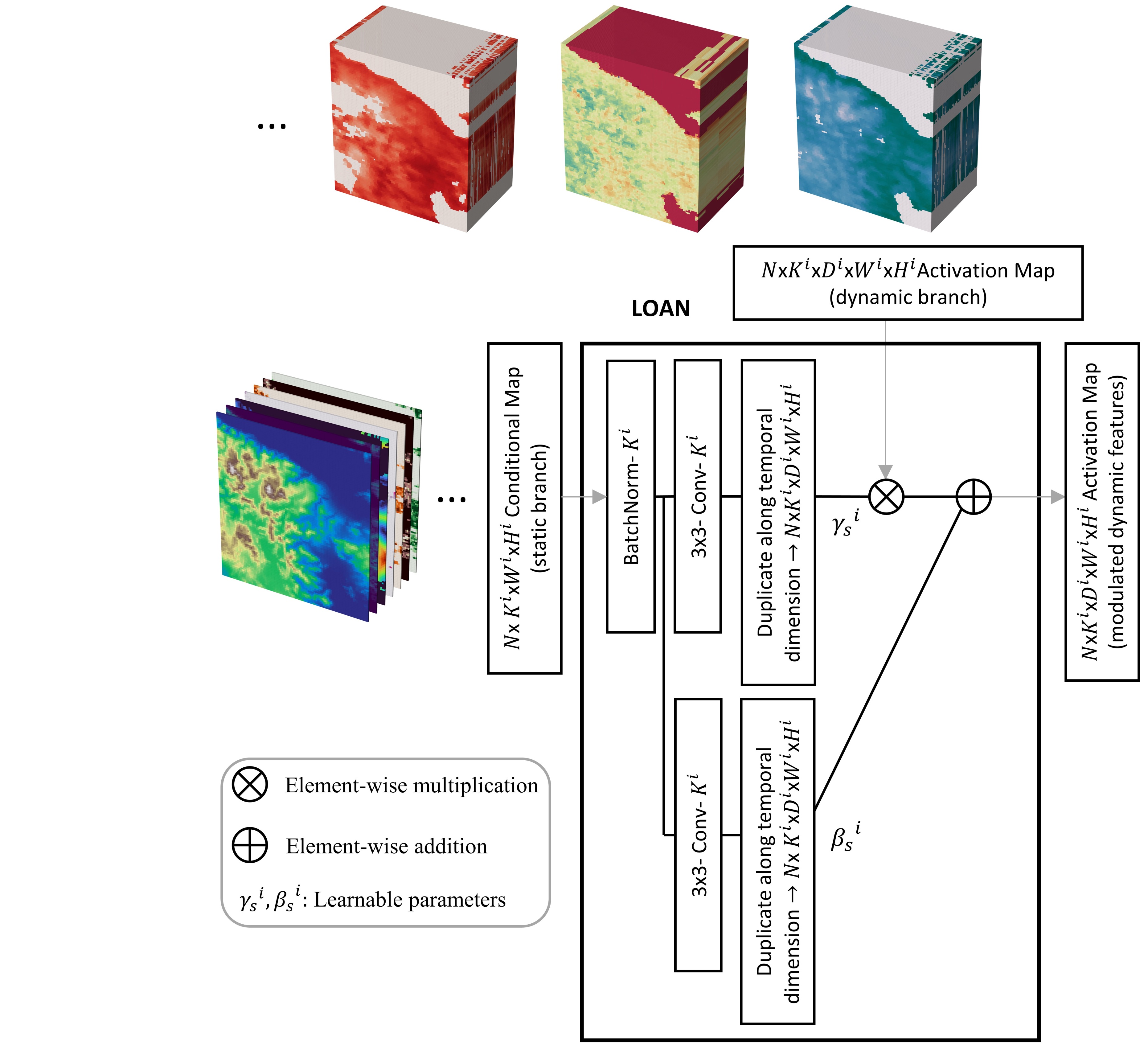}
    \caption{Illustration of the Location-aware Adaptive Normalization layer (LOAN) when using the conditional map from the static branch. The conditional map of the static branch and the activation map of the dynamic branch have the same $N$, $K^i$, $W^i$, and $H^i$ dimensions. BatchNorm denotes a batch normalization layer. $3{\times}3$-Conv-$K^i$ denotes a convolution layer with a kernel size of $3 \times 3$ and $K^i$ output channels.}
    \label{fig:2}
\end{figure}

The way how ${\gamma_s}_{(n,k,w,h)}^i$ and ${\beta_s}_{(n,k,w,h)}^i$ are computed is illustrated in Fig.~\ref{fig:2}. First, the conditional map ${z_s}_{}^{i}$ is normalized channel-wise as following:
\begin{equation}
    \label{eq:5}
    {\hat{z_s}}_{(n,k,w,h)}^i =\frac{{z_s}_{(n,k,w,h)}^i-{\mu_k}_{}^i}{{\sigma_k}_{}^i}\,,
\end{equation}
where
\begin{align}
    {\mu_k}_{}^i = \frac{1}{N W^i H^i}\sum_{n=1}^N\sum_{w=1}^{W^i}\sum_{h=1}^{H^i}{z_s}_{(n,k,w,h)}^{i} \label{eq:6}\,,\\
    {\sigma_k}_{}^i = \sqrt{\frac{1}{N W^i H^i}\sum_{n=1}^N\sum_{w=1}^{W^i}\sum_{h=1}^{H^i}{({z_s}_{(n,k,w,h)}^{i}{-}{\mu_k}_{}^{i})}^{2}}\,. \label{eq:7}
\end{align}

Afterward, ${\hat{z_s}}_{(n,k,w,h)}^i$ is projected by two convolutional layers with $K^i$ filters to compute ${\gamma_s}_{(n,k,w,h)}^i$ and ${\beta_s}_{(n,k,w,h)}^i$. In our implementation, these modulation parameters are then duplicated along the temporal dimension to match the depth $D^i$ of  ${z_d}_{}^i$ such that \eqref{eq:8} can be computed.

We add the Location-aware Adaptive Normalization layer (LOAN) in the first two blocks as shown in Fig.~\ref{fig:1}. The activation maps of the dynamic branch are normalized only in the $1^{\text{st}}$ block and modulated in both the $1^{\text{st}}$ and $2^{\text{nd}}$ blocks. The impact of the blocks where the LOAN layer is added is evaluated in Table~\ref{tab:table6}.

\begin{figure}[t]
    \centering
    \includegraphics[width=3.5in]{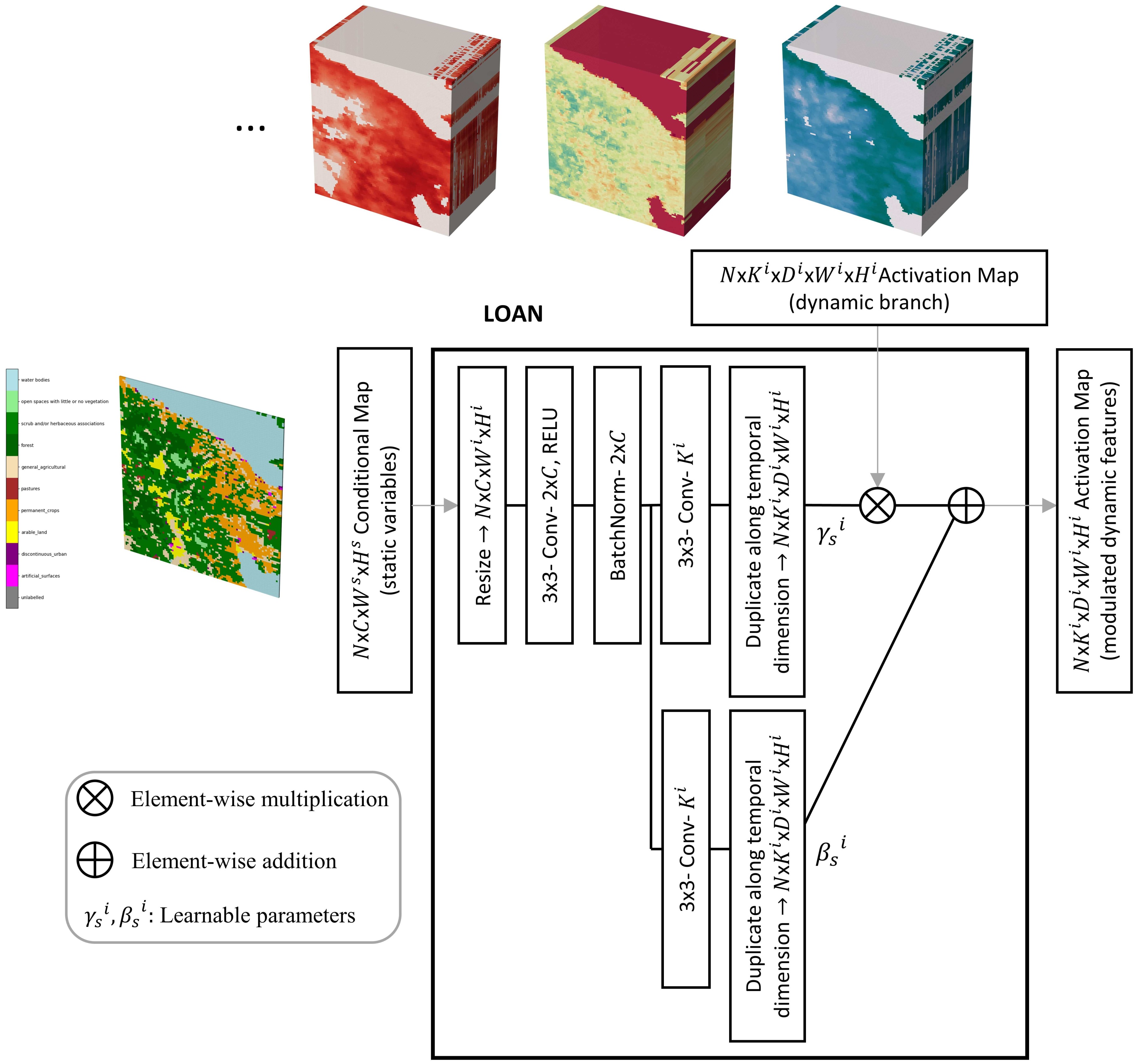}
    \caption{
    Illustration of the Location-aware Adaptive Normalization layer (LOAN) when using $C$ static variables directly for modulation. In this case, there is a mismatch between the dimensions of the conditional map and the activation maps of the dynamic branch at the $i\text{-th}$ block. To adjust for the spatial resolution, we resize the conditional map using nearest-neighbor down-sampling to match the resolution of the activation map from the dynamic branch. The convolution layer takes $C$ channels as input and generates an output with $K^i$ channels.         
    }
    \label{fig:3}
\end{figure}

In the experimental section, we also evaluate a variant of LOAN that is not conditioned on the intermediate features of the static branch as shown in Figs.~\ref{fig:1} and \ref{fig:2}, but on the static variables directly as shown in Fig.~\ref{fig:3}. In this case, the conditional map has different spatial dimensions and number of channels compared to the features in the dynamic branch, i.e., the conditional maps consist of $C$ variables and have $W^s \times H^s$ spatial dimensions. In this respect, the conditional map ($W^s \times H^s$) is first resized to match the spatial dimensions ($W^i \times H^i$) of the activation map from the dynamic branch. We use the nearest-neighbor method for the down-sampling. The resized conditional map is then fed into a convolutional layer with $3 \times 3$ kernel size to double the number of channels, i.e., $2 \times C$. Finally, as in the previous version of LOAN, the conditional map is normalized, projected by two convolutional layers, and duplicated along the temporal dimension to compute ${\gamma_s}_{(n,k,w,h)}^i$ and ${\beta_s}_{(n,k,w,h)}^i$. The impact of different conditional maps using the variants of LOAN is evaluated in Table \ref{tab:table4}.

\subsection{Absolute Temporal Encoding (TE)}\label{sec:IV-C}
\begin{figure}[h]
    \centering
    \includegraphics[width=3.4in]{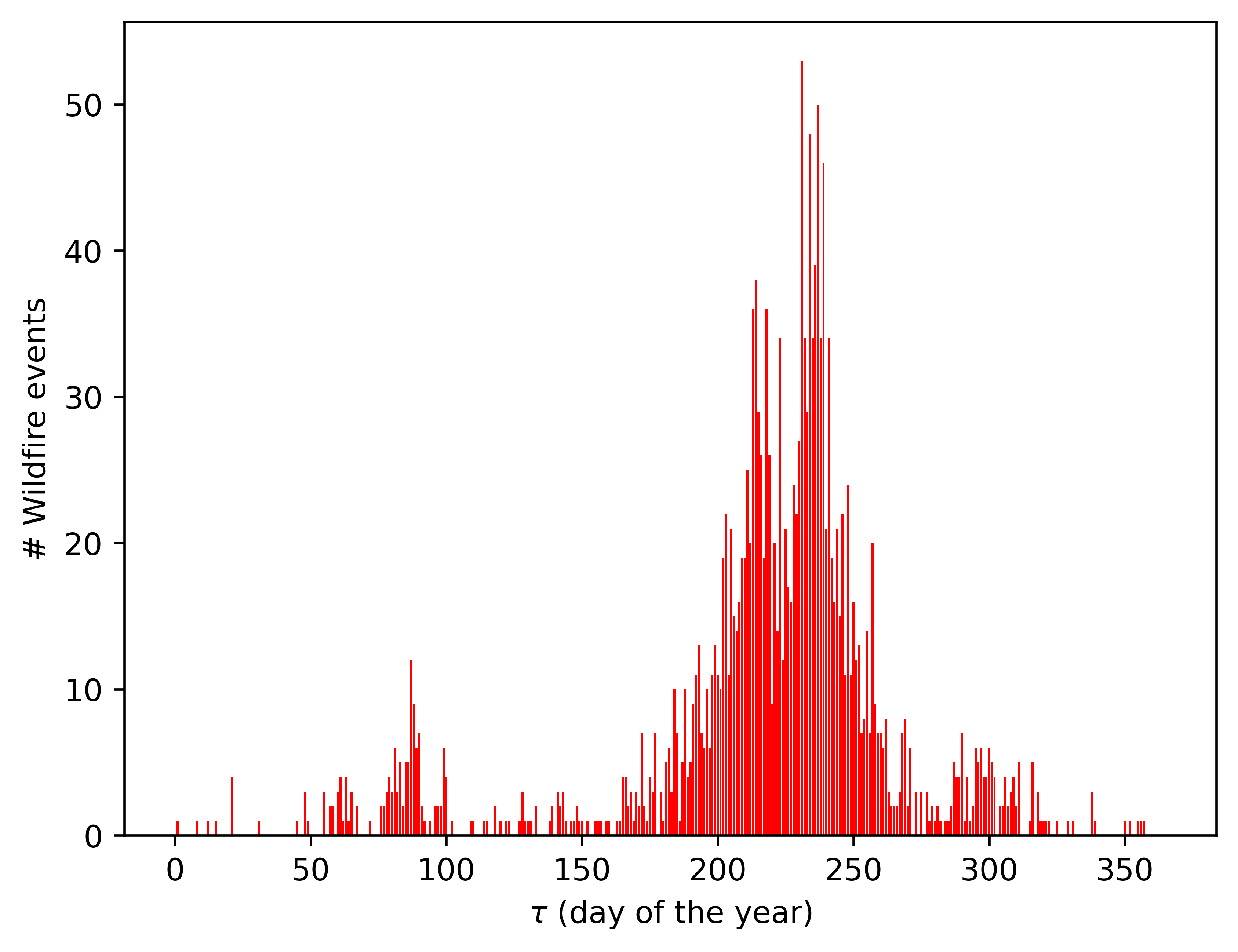}
    \caption{Distribution of the wildfire events per day between $06/03/2009$ and $29/08/2021$ (FireCube dataset).}
    \label{fig:5}
\end{figure}

Some extreme events in the climate model have a dependent relation on time \cite{doi:10.1080/19475705.2018.1526219, hantson2016status}. This is also the case for the FireCube dataset \cite{prapas_ioannis_2022_6475592} where wildfire events vary from month to month and occur more frequently in the summer time as shown in Fig.~\ref{fig:5}. So far, the network does not consider an absolute time like the day of the year. Instead, for any forecast day $T{+}1$, the last 10 days are used as observations but the network does not have the information what day during the year $T$ is.  

As shown in Fig.~\ref{fig:1}, we add this information to the dynamic branch before we concatenate the static and dynamic features. To encode the day of the year, we use the standard fixed sinusoidal-based encoding by Vaswani et al.~\cite{vaswani2017attention}. We pre-compute for each day of the year $\tau{\in}\left[0,365\right]$\footnote{We consider February $29$ for the encoding.}, which is extracted from $T$, the absolute temporal encoding vector $\textbf{X}_\tau \in \mathbb{R}^{256}$:  
\begin{align}
    \textbf{X}_\tau(2j)\ =\ \sin(\tau/{10}^{2j/256})\,, \label{eq:11}\\
    \textbf{X}_\tau(2j{+}1)\ =\ \cos{(\tau/{10}^{2j/256})}\,, \label{eq:12}
\end{align}
where $j$ is the embedding dimension. Each even dimension results from a sine function, while each odd dimension results from a cosine function. This allows $\tau$ to have a smooth and yet unique encoding for every time step, i.e., each day of a year. Note that the vector has the same size as $\textbf{X}_d$. 

\begin{figure}[t]
    \centering
    \includegraphics[width=3.45in]{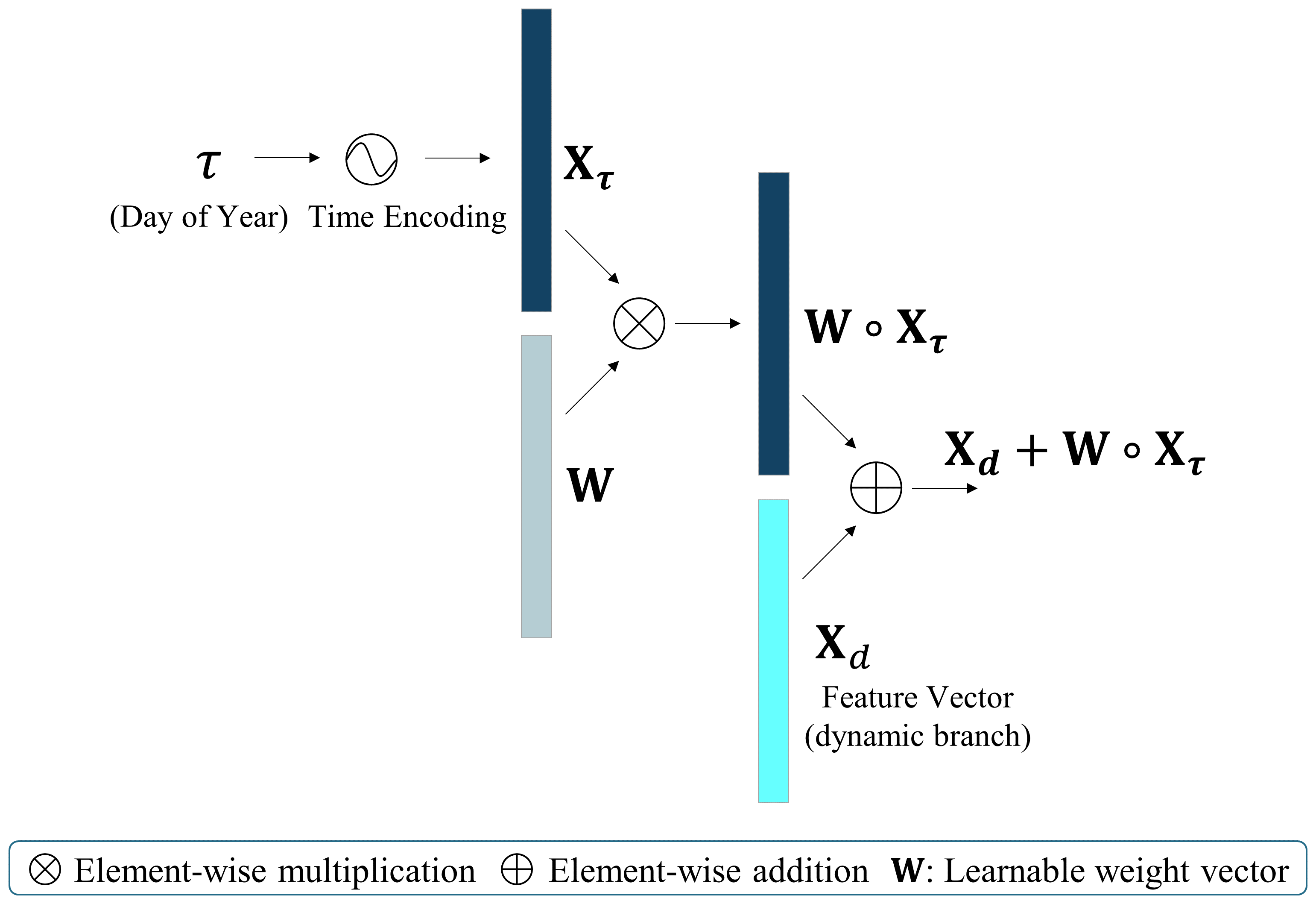}
    \caption{Illustration of how the absolute temporal encoding is added in the model. The day of year $\tau$ is encoded into a vector $\textbf{X}_\tau$ and each element is weighted by the learned weight vector $\textbf{W}$. The weighted vector is then added to the dynamic feature vector $\textbf{X}_d$.}
    \label{fig:4}    
\end{figure}

In order to add the absolute time encoding vector $\textbf{X}_\tau$ to the dynamic feature vector $\textbf{X}_d$, we weight each element of the vector by a learnable weight vector $\textbf{W}\in\mathbb{R}^{256}$:     
\begin{equation}
    \textbf{X}_d + \textbf{W}\circ\textbf{X}_\tau\,,
    \label{eq:10}
\end{equation}
where $\circ$ denotes the Hadamard product, i.e., element-wise multiplication. Fig.~\ref{fig:4} illustrates how the temporal embedding is added to the model. 

\afterpage{
\begin{center}
\begin{table*}[h!]
\caption{Comparison with baselines. The classification metrics are shown for the years $2019$-$2021$. The values of Precision, Recall, F1-score, OA, and AUROC are given in percent (\%). (TE) denotes the absolute temporal encoding. \label{tab:table1}}
\centering
\setlength\tabcolsep{4pt}
\begin{tabular}{r *{9}c}     
\toprule
\multicolumn{10}{c}{Year $2019$ (val) - $6430$ samples} \\
\midrule
Algorithm & TP($\uparrow$) & FP($\downarrow$) & TN($\uparrow$) & FN($\downarrow$) & Precision($\uparrow$) & Recall($\uparrow$) & F1-score($\uparrow$) & AUROC($\uparrow$) & OA($\uparrow$)\\
\midrule
RF \cite{Prapas2} & 575 & 372 & 4758 & 725 & 60.72 & 44.23 & 51.18 & 88.54 & 82.94 \\
XGBoost \cite{Prapas2} & 928 & 448 & 4682 & 372 & 67.44 & 71.38 & 69.36 & 92.33 & 87.25 \\
LSTM \cite{Prapas2} & 968 & 431 & 4699 & 332 & 69.19 & 74.46 & 71.73 & 93.63 & 88.13 \\
ConvLSTM \cite{Prapas2} & 867 & 276 & 4854 & 433 & 75.85 & 66.69 & 70.98 & 94.69 & 88.97 \\
TimeSformer \cite{gberta_2021_ICML} & 967 & \underline{248} & \underline{4882} & 333 & 79.59 & 74.38 & \underline{76.90} & \textbf{95.71} & 90.96 \\
SwinTransformer \cite{Liu_2022_CVPR} & \textbf{979} & 324 & 4806 & \textbf{321} & 75.13 & \textbf{75.31} & 75.22 & 94.54 & 89.97 \\
3D CNN & 918 & 265 & 4865 & 382 & 77.60 & 70.62 & 73.94 & 94.17 & 89.94 \\ 
\midrule
\textbf{2D/3D CNN} & \underline{970} & \underline{248} & \underline{4882} & \underline{330} & \underline{79.64} & \underline{74.62} & \textbf{77.05} & 94.52 & \underline{91.01} \\
\textbf{2D/3D CNN w/ TE} & 905 & \textbf{182} & \textbf{4948} & 395 & \textbf{83.26} & 69.62 & 75.83 & \underline{95.08} & \textbf{91.03} \\
\midrule
\midrule
\multicolumn{10}{c}{Year $2020$ (test) - $6088$ samples} \\
\midrule
Algorithm & TP($\uparrow$) & FP($\downarrow$) & TN($\uparrow$) & FN($\downarrow$) & Precision($\uparrow$) & Recall($\uparrow$) & F1-score($\uparrow$) & AUROC($\uparrow$) & OA($\uparrow$)\\
\midrule
RF \cite{Prapas2} & 750 & 245 & 4615 & 478 & 75.38 & 61.07 & 67.48 & 91.17 & 88.12 \\
XGBoost \cite{Prapas2} & \textbf{891} & 322 & 4538 & \textbf{337} & 73.45 & \textbf{72.56} & 73.00 & 91.12 & 89.18 \\
LSTM \cite{Prapas2} & \textbf{891} & 290 & 4570 & \textbf{337} & 75.44 & \textbf{72.56} & \underline{73.97} & 93.60 & 89.70 \\
ConvLSTM \cite{Prapas2} & 811 & 155 & 4705 & 417 & 83.95 & 66.04 & 73.93 & \textbf{94.31} & 90.60 \\
TimeSformer \cite{gberta_2021_ICML} & 751 & \underline{140} & \underline{4720} & 477 & \underline{84.29} & 61.16 & 70.88 & 92.41 & 89.87 \\
SwinTransformer \cite{Liu_2022_CVPR} & 794 & 202 & 4658 & 434 & 79.72 & 64.66 & 71.40 & 92.79 & 89.55 \\
3D CNN & 797 & 160 & 4700 & 431 & 83.28 & 64.90 & 72.95 & 93.10 & 90.29 \\
\midrule
\textbf{2D/3D CNN} & \underline{841} & 160 & 4700 & \underline{387} & 84.02 & \underline{68.49} & \textbf{75.46} & 93.98 & \textbf{91.02} \\
\textbf{2D/3D CNN w/ TE} & 776 & \textbf{117} & \textbf{4743} & 452 & \textbf{86.90} & 63.19 & 73.17 & \underline{94.20} & \underline{90.65} \\
\midrule
\midrule
\multicolumn{10}{c}{Year $2021$ (test) - $36732$ samples} \\
\midrule
Algorithm & TP($\uparrow$) & FP($\downarrow$) & TN($\uparrow$) & FN($\downarrow$) & Precision($\uparrow$) & Recall($\uparrow$) & F1-score($\uparrow$) & AUROC($\uparrow$) & OA($\uparrow$)\\
\midrule
RF \cite{Prapas2} & 3264 & 1157 & 31168 & 1143 & 73.83 & 74.06 & 73.95 & 96.82 & 93.74 \\
XGBoost \cite{Prapas2} & 3016 & 1345 & 30980 & 1391 & 69.16 & 68.44 & 68.80 & 95.88 & 92.55 \\
LSTM \cite{Prapas2} & 3739 & 1359 & 30966 & 668 & 73.34 & 84.84 & 78.67 & 97.13 & 94.48 \\
ConvLSTM \cite{Prapas2} & 3514 & 769 & 31556 & 893 & 82.05 & 79.74 & 80.87 & 97.76 & 95.48 \\
TimeSformer \cite{gberta_2021_ICML} & 3578 & 867 & 31458 & 829 & 80.49 & 81.19 & 80.84 & 97.67 & 95.38 \\
SwinTransformer \cite{Liu_2022_CVPR} & \textbf{3962} & 954 & 31371 & \textbf{445} & 80.59 & \textbf{89.90} & 84.99 & 98.09 & 96.19 \\
3D CNN & 3766 & 810 & 31515 & 641 & 82.30 & 85.45 & 83.85 & 98.02 & 96.05 \\
\midrule
\textbf{2D/3D CNN} & \underline{3870} & \underline{757} & \underline{31568} & \underline{537} & \underline{83.64} & \underline{87.81} & \underline{85.68} & \underline{98.19} & \underline{96.48} \\
\textbf{2D/3D CNN w/ TE} & 3841 & \textbf{416} & \textbf{31909} & 566 & \textbf{90.23} & 87.16 & \textbf{88.67} & \textbf{98.54} & \textbf{97.33} \\
\bottomrule
\end{tabular}
\end{table*}
\end{center}

\begin{table*}[h!]
\begin{center}
\caption{Quantitative results of different deep learning model designs. The classification metrics are given in percent (\%). Additionally, the total number of parameters (\# Params), estimated multiply-accumulate operations given in Mega (MMACs),} and the inference time as samples per millisecond (\# SPmS) are provided. (TE) denotes the absolute temporal encoding and (LOAN) the location-aware adaptive normalization layer. \label{tab:table2}
\setlength\tabcolsep{3pt}
\setlength\extrarowheight{1pt}
\begin{tabular}{r *{11}c}
\toprule
\multicolumn{11}{c}{Year $2020$-$2021$ (test)} \\
\midrule
Algorithm & LOAN & TE & \# Params($\downarrow$) & MMACs ($\downarrow$) & \# SPmS($\uparrow$) & Precision($\uparrow$) & Recall($\uparrow$) & F1-score($\uparrow$) & AUROC($\uparrow$) & OA($\uparrow$) \\
\midrule
LSTM \cite{Prapas2} & $\times$ & $\times$ & 30k & 0.27 & 955$\pm$43 & 73.74 & 82.17 & 77.72 & 96.53 & 93.80 \\
\midrule
ConvLSTM \cite{Prapas2}  & $\times$ & $\times$ & 372k & 417.08 & 7$\pm$0 & 82.40 & 76.75 & 79.47 & 97.12 & 94.78 \\
\midrule
TimeSformer \cite{gberta_2021_ICML} & $\times$ & $\times$ & 1.16m & 831,667.58 & 2$\pm$0 & 81.13 & 76.82 & 78.92 & 96.57 & 94.60 \\
\midrule
SwinTransformer \cite{Liu_2022_CVPR} & $\times$ & $\times$ & 1.78m & 122,346.56 & 1$\pm$0 & 80.45 & \textbf{84.40} & 82.38 & 97.06 & 95.25 \\
\midrule
\multirow{2.5}{*}{3D CNN} & $\times$ & $\times$ & 323k & 476.56 & 18$\pm$1 & \underline{83.93} & 77.48 & 80.58 & 97.15 & 95.08 \\
\cmidrule{2-11}
 & $\times$ & $\times$ & 499k & 585.11 & 17$\pm$1 & 82.47 & 80.98 & 81.72 & 97.15 & 95.23 \\
\midrule
\multirow{5.5}{*}{\textbf{2D/3D CNN}} & $\times$ & $\times$ & 321k & 137.24 & 47$\pm$0 & 83.35 & 79.98 & 81.63 & 97.11 & 95.26\\
\cmidrule{2-11}
 & $\surd$ & $\times$ & 413k & 168.23 & 33$\pm$1 & 83.71 & 83.60 & 83.65 & 97.41 & 95.70 \\
\cmidrule{2-11}
 & $\times$ & $\surd$ & 321k & 137.24 & 47$\pm$1 & 83.90 & \underline{84.22} & \underline{84.06} & \underline{97.64} & \underline{95.80}\\
\cmidrule{2-11}
 & $\surd$ & $\surd$ & 414k & 168.23 & 33$\pm$2 & \textbf{89.65} & 81.93 & \textbf{85.62} &\textbf{97.78} & \textbf{96.38} \\
\bottomrule
\end{tabular}
\end{center}
\end{table*}
}
\section{Experimental Results and Analysis}
\label{sec:V}
\emph{Implementation Details}.
The network is trained with the binary cross entropy loss \eqref{eq:13} using the Pytorch framework \cite{NEURIPS2019_9015} with a learning rate $0.00003$ and the Adam optimizer ($\beta_1=0.9,~ \beta_2=0.999$) \cite{Kingma2015AdamAM} with a weight decay $0.02$. We use a batch size of $N=256$ and train the network for $40$ epochs. All models were trained on a single NVIDIA GeForce RTX 3090 GPU.

\emph{Performance Metrics}. As described in Section~\ref{sec:III}, we use the FireCube dataset \cite{prapas_ioannis_2022_6475592}. We follow the same protocol for quantitative comparison as in Prapas et al.~\cite{Prapas1} and Kondylatos et al.~\cite{Prapas2}. 
The evaluation metrics are precision, recall, and F1-score, calculated for the positive class that represents a wildfire event. 
In addition, we report true positives (TP), false positives (FP), true negatives (TN), and false negatives (FN). Moreover, we provide the overall accuracy (OA) and the area under the receiver operating characteristic curve (AUROC) as evaluation metrics. OA is the accuracy obtained on all negative and positive samples in the test set. The AUROC describes the true positive rate (TPR) against the false positive rate (FPR) within multiple thresholds in one value.

\subsection{Comparison with Baselines}\label{sec:V-A}

We compare our approach to the approaches that have been evaluated on the described dataset in Kondylatos et al.~\cite{Prapas2}. This includes two deep learning models, namely LSTM \cite{LSTM} and ConvLSTM \cite{shi2015convolutional}, and two classical machine learning classifiers, namely Random Forests (RF) \cite{Breiman2001} and XGBoost \cite{XGBoost}. For further details regarding the architectures and hyper-parameters of the models, we refer to the work of Kondylatos et al.~\cite{Prapas2}. In order to demonstrate the benefit of treating static and dynamic variables differently, we also compare with a one-branch 3D CNN where we duplicate the static variables along the temporal dimension and concatenate them together with the dynamic variables to form a single data cube. 
In addition, we compare to the recent transformer models TimeSformer \cite{gberta_2021_ICML} and Video SwinTransformer \cite{Liu_2022_CVPR}, which use space-time attention. As mentioned in Subsection~\ref{sec:II-D}, vision transformers are based on a self-attention mechanism to model spatio-temporal dependencies. Compared to CNNs, transformers have less inductive bias and need much more computational resources for training. To the best of our knowledge, no prior work has done a systematic study about the performance of video vision transformers for wildfire forecasting.

To ensure a fair comparison, all baselines were re-implemented and trained on the same samples with a fixed random seed. We do not use any augmentation technique. The quantitative results of our experiments are provided in Table \ref{tab:table1}. The results of the proposed 2D/3D CNN are shown with and without absolute temporal encoding (TE). 

We can observe that the proposed 2D/3D CNN outperforms the other methods for most metrics, particularly FP, TN, Precision, F1-score, and OA, on the validation and testing sets. SwinTransformer and ConvLSTM achieve a slightly higher AUROC for 2019 and 2020, respectively. LSTM and SwinTransformer achieve a higher recall, but at the cost of a very low precision. In comparison with other deep learning methods, LSTM and SwinTransformer have even the highest number of false positives for all years. The main weakness of LSTM lies in the fact that it does not consider the spatial context, while large models like SwinTransformer are prone to overfitting, which results in a relatively poor performance for 2020.
RF and XGBoost have the same disadvantage as LSTM, but even a weaker temporal model and thus perform worse than LSTM.   
Most interesting is the comparison to 3D CNN since it uses the same 3D CNN structure but only one branch, i.e., it treats static variables like dynamic variables. The results show that the proposed approach with two branches outperforms the single-branch architecture for all metrics and all years. This demonstrates the importance of treating static variables differently than dynamic variables. 
Adding the absolute temporal encoding (TE) to the model substantially reduces FP at the cost of decreasing TP. This is also reflected in the precision and recall.

\begin{table*}[!h]
\begin{center}
\caption{Performance metrics for different categories of static variables. The values are given in percent (\%). \label{tab:table3}}
\setlength\tabcolsep{3.4pt}

\begin{tabular}{l *{12}c}

\toprule
 & \multicolumn{5}{c}{Year $2019$ (val)} & ~~ & ~~ & \multicolumn{5}{c}{Year $2020$-$2021$ (test)}\\
\midrule
Static Variables & Precision($\uparrow$) & Recall($\uparrow$) & F1-score($\uparrow$) & AUROC($\uparrow$) & OA($\uparrow$) & & & Precision($\uparrow$) & Recall($\uparrow$) & F1-score($\uparrow$) & AUROC($\uparrow$) & OA($\uparrow$) \\
\midrule
DEM + slope & 75.61 & 73.69 & 74.64 & 93.82 & 89.88 & & & 80.88 & 81.67 & 81.27 & 97.06 & 95.05 \\
\midrule
\parbox{2.5cm}{Distance to roads \\ Distance to waterway \\ Population density} & 77.27 & 76.08 & 76.67 & 94.16 & 90.63 & & & 81.91 & 74.48 & 78.02 & 96.11 & 94.48 \\
\midrule
Land cover & 77.27 & 76.62 & 76.94 & 94.62 & 90.72 & & & 81.85 & 79.93 & 80.88 & 96.92 & 95.03 \\
\midrule
\midrule
\parbox{2.5cm}{DEM + slope \\ Distance to roads \\ Distance to waterway \\ Population density} & 75.09 & \textbf{77.00} & 76.03 & 94.09 & 90.19 & & & 79.64 & 76.89 & 78.24 & 96.69 & 94.37 \\
\midrule
\parbox{2.5cm}{DEM + slope \\ Land cover} & 74.61 & 73.46 & 74.03 & 94.24 & 89.58 & & & 80.05 & 81.44 & 80.74 & 97.20 & 94.89 \\
\midrule
\parbox{2.5cm}{Land cover \\ Distance to roads \\ Distance to waterway \\ Population density} & 77.52 & 73.23 & 75.32 & \textbf{94.52} & 90.30 & & & 82.63 & 81.12 & 81.87 & 96.98 & 95.27 \\
\midrule
\midrule
\textbf{All static variables} & \textbf{79.64} & 74.62 & \textbf{77.05} & \textbf{94.52} & \textbf{91.01} & & & \textbf{83.71} & \textbf{83.60} & \textbf{83.65} & \textbf{97.41} & \textbf{95.70} \\
\bottomrule
\end{tabular}
\end{center}
\end{table*}

\begin{table*}[t]
\begin{center}
\caption{Impact of different conditional maps on the feature modulation. The evaluation metrics are given in percent (\%). \label{tab:table4}}
\setlength\tabcolsep{3.4pt}
\begin{tabular}{l *{12}c}
\toprule
 & \multicolumn{5}{c}{Year $2019$ (val)} & ~~ & ~~ & \multicolumn{5}{c}{Year $2020$-$2021$ (test)}\\
\midrule
Conditional Map & Precision($\uparrow$) & Recall($\uparrow$) & F1-score($\uparrow$) & AUROC($\uparrow$) & OA($\uparrow$) & & & Precision($\uparrow$) & Recall($\uparrow$) & F1-score($\uparrow$) & AUROC($\uparrow$) & OA($\uparrow$) \\
\midrule
$~~~~~~~~-$ & 75.68 & 62.23 & 68.30 & 93.38 & 88.32 & & & 83.35 & 79.98 & 81.63 & 97.11 & 95.26 \\
\midrule
DEM + slope & 70.84 & 71.00 & 70.92 & 92.77 & 88.23 & & & 76.90 & 79.45 & 78.15 & 96.16 & 94.15 \\
\midrule
\parbox{2.5cm}{Distance to roads \\ Distance to waterway \\ Population density} & 73.83 & \textbf{76.15} & 74.97 & 93.74 & 89.72 & & & 77.98 & 76.63 & 77.30 & 96.22 & 94.08 \\
\midrule
Land cover & 79.05 & 72.00 & 75.36 & 94.58 & 90.48 & & & 81.53 & 75.19 & 78.23 & 96.80 & 94.49 \\
\midrule
All static variables & 77.12 & 74.92 & 76.00 & \textbf{94.95} & 90.44 & & & 81.05 & \textbf{84.08} & 82.54 & \textbf{97.55} & 95.32\\
\midrule
\parbox{2cm}{\textbf{Activation maps \\ (Static branch)}} & \textbf{79.64} & 74.62 & \textbf{77.05} & 94.52 & \textbf{91.01} & & & \textbf{83.71} & 83.60 & \textbf{83.65} & 97.41 & \textbf{95.70} \\
\bottomrule
\end{tabular}
\end{center}
\end{table*}

In Table \ref{tab:table2}, we present additional experimental results alongside the memory footprint as the number of parameters (\# Params), the estimated multiply-accumulate operations (multiply-adds) (MMACs), and the expected inference time, which is estimated as samples per millisecond (\# SPmS). The performance metrics are calculated on both testing years $2020$-$2021$ as one set. Since the LOAN layers increase the amount of parameters of the 2D/3D CNN, we report the results of the one-branch 3D CNN using $323$k and $499$k parameters. The smaller 3D CNN has about the same amount of parameters as the 2D/3D CNN without LOAN, whereas the larger 3D CNN has more parameters than the proposed model. Due to the lack of a spatial modeling, LSTM has the fewest parameters and is the fastest, but the precision is very low. ConvLSTM, the small one-branch 3D CNN, and 2D/3D CNN without LOAN and TE, which consider the spatial context, perform similar but 2D/3D CNN is the fastest approach and ConvLSTM is the slowest approach among them. Compared to the other approaches, transformer models have considerably more parameters and require more operations, which makes them computationally expensive. Nevertheless, our approach outperforms both transformer models while requiring by far less parameters and computational operations.

Normalizing the dynamic features conditioned on the static features (LOAN) increases all metrics. It also outperforms the large 3D CNN in all metrics and inference time. Adding TE increases all metrics when LOAN is not used, while increasing the computational cost only very little. When LOAN is used, adding TE decreases the recall but increases all other metrics. Since LOAN and TE change the dynamic features, we observe a different trade-off between recall and precision if both are used. This change is consistent over the years as shown in Table \ref{tab:table1}. Nevertheless, the F1-Score, AUROC, and OA are highest if both are used.                         

\subsection{Variable Importance} \label{sec:V-B}

To assess the importance of different static variables, we present the results obtained with different combinations of static variables in Table \ref{tab:table3}. For this experiment, we use 2D/3D CNN with LOAN but without TE. All dynamic variables are used in this experiment and the results are reported for the year $2019$ and for the years $2020$-$2021$ as one set. The static variables are grouped into 3 main categories: topographic variables consisting of digital elevation model (DEM) and slope, anthropogenic-related variables consisting of distance to roads or waterway and population density, and land cover variables. 

From the results in Table \ref{tab:table3}, we can conclude that among the 3 categories topographic variables give the best results for the years $2020$-$2021$ when they are used without other variables. While land cover and anthropogenic-related variables provide the best results for the year $2019$. Overall, all static variables are relevant and the best results are obtained when all static variables are used (last row). This is also better than using the static variables of 2 out of the 3 categories, which is reported in rows 4-6 of Table \ref{tab:table3}.      

\subsection{Comparing Different Conditional Maps} \label{sec:V-C}
While Table \ref{tab:table3} shows the importance of different static variables as input to the 2D/3D CNN, we also analyze the impact of different ways to modulate the dynamic features in Table \ref{tab:table4}. For the experiments, we use the 2D/3D CNN without TE and all dynamic and static variables as input. While we use all variables, the different settings differ in the input that is fed to the LOAN layer, i.e., the static features that the modulation of the dynamic features is conditioned on. 

The results of the proposed conditioning, where we use the features from the corresponding block of the static branch, are shown in the last row. In the first row, we show the results if we do not use LOAN at all, i.e., we do not use any modulation of the dynamic features.                  
For the other rows of Table~\ref{tab:table4}, we modify LOAN such that it is not conditioned on the intermediate features of the static branch as shown in Figs.~\ref{fig:1} and \ref{fig:2}. Instead, we condition LOAN directly on static variables. Note that we need to slightly adapt LOAN as shown in Fig.~\ref{fig:3} since the number of static input variables $C$ differs from the number of feature channels $K$ at the block where LOAN is added.     

As seen from Table \ref{tab:table4}, the best result is obtained when we use the activation maps, i.e., the intermediate features, from the static branch for conditioning the modulation. However, comparable results are obtained when all static variables are directly used by the variant of LOAN that is shown in Fig.~\ref{fig:4}. While this variant achieves slightly higher AUROC, the variant shown in Fig.~\ref{fig:3} achieves higher F1-score and OA. Using the static variables directly, we can analyze how the three categories of static variables impact the modulation of the dynamic features and thus the results. We can conclude that the modulation of the dynamic features is very sensitive to its condition. For the year $2019$, all three categories (rows $2$-$4$) improve the results compared to the setup without feature modulation (first row). For the years $2020$-$2021$, this is not the case and only the combination of all static variables (row $5$) leads to an improvement. The reason is the mismatch between $C$ and $K^i$. Compared to the other categories, land cover has the highest number of variables per pixels ($C{=}10$) and shows the best performance. This indicates that conditioning the modulation of the dynamic features on the intermediate features of the static branch is a more practical approach than conditioning the modulation on the static variables directly, which seems to be sensitive to the number of variables.

\subsection{Qualitative Results}\label{sec:V-D}
\begin{figure*}[p]
    \centering
    \includegraphics[width=0.99\textwidth]{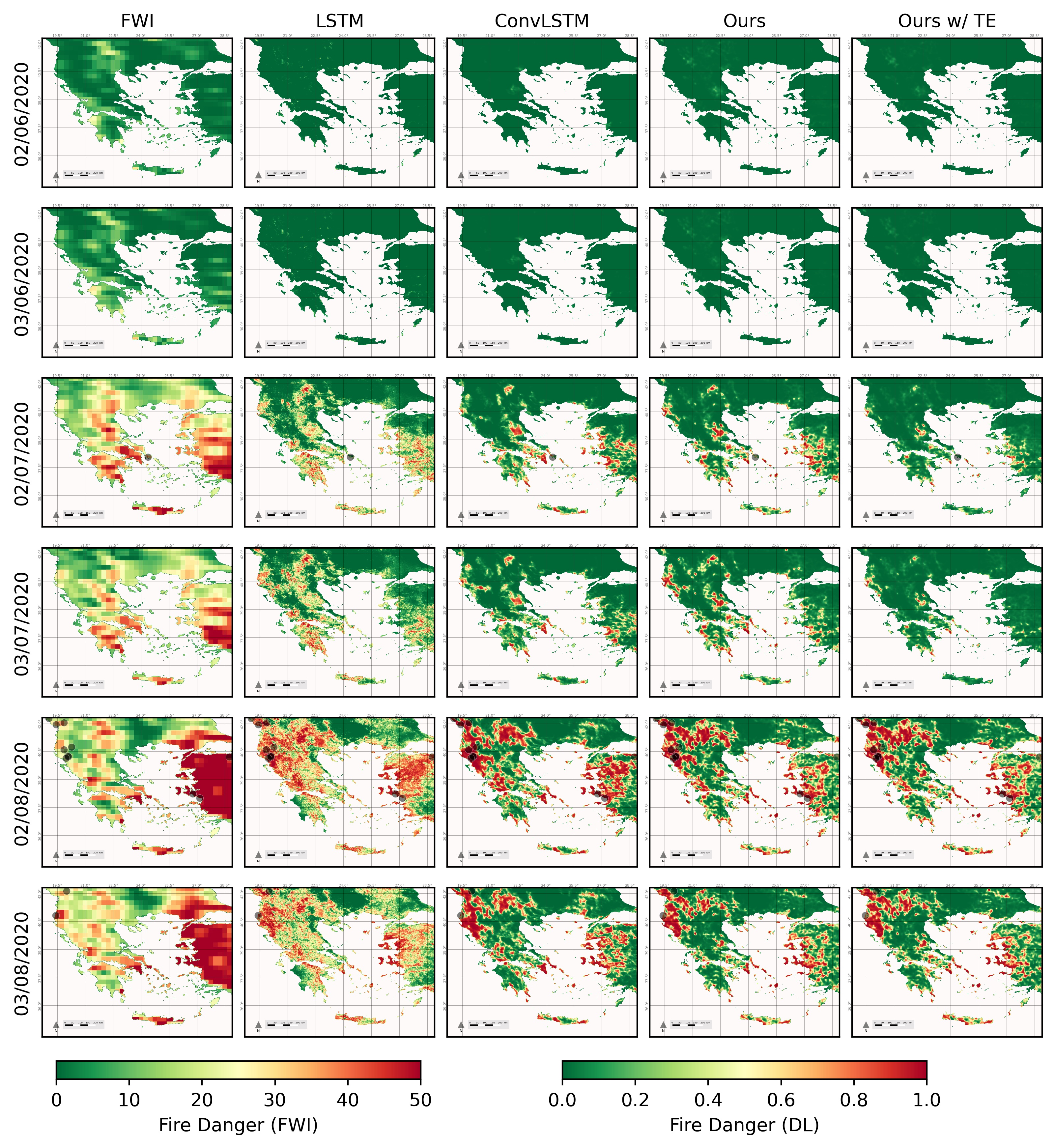}
    \caption{Qualitative results for $6$ days during the wildfire season in year $2020$. The black circles represent an ignition of a large wildfire on that day. (TE) denotes absolute temporal encoding.}
    \label{fig:6}
\end{figure*}

\begin{figure*}[p]
    \centering
    \includegraphics[width=0.99\textwidth]{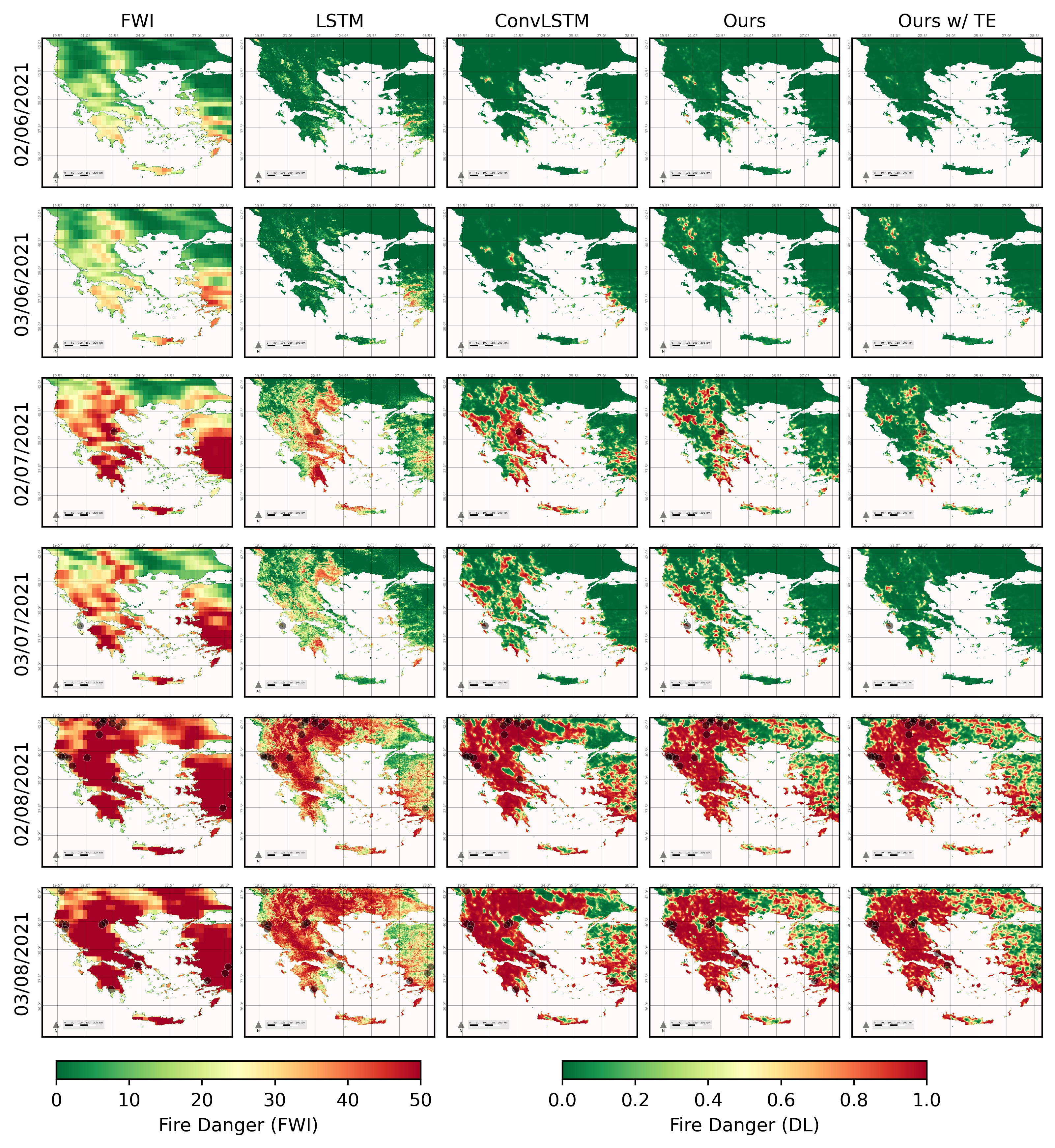}
    \caption{Qualitative results for $6$ days during the wildfire season in the extreme year $2021$. The black circles represent an ignition of a large wildfire on that day. (TE) denotes absolute temporal encoding.}
    \label{fig:7}
\end{figure*}

\begin{figure*}[!t]
  \centering
  \includegraphics[width=0.99\textwidth]{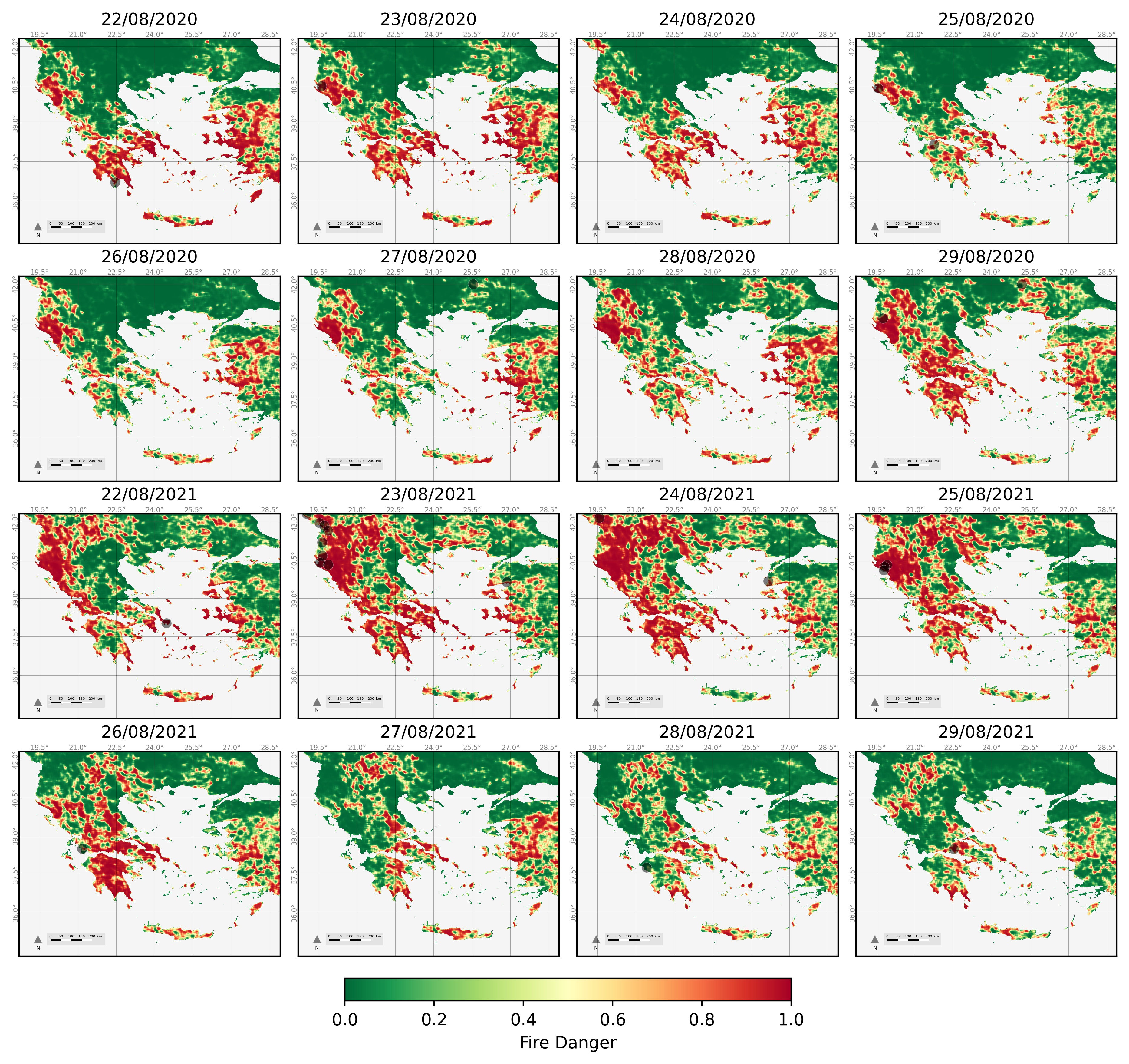}
  \caption{Predictions for $8$ consecutive days in August. The upper two rows show the results for the year $2020$ and the last two rows show the results for the extreme year $2021$. The maps are produced by the proposed 2D/3D CNN with absolute temporal encoding. The black circles represent an ignition of a large wildfire on that day.}
  \label{fig:8}
\end{figure*}

Predicted wildfire danger-susceptibility maps are depicted in Figs.~\ref{fig:6}, \ref{fig:7}, and \ref{fig:8}. We take input from the $1^{st}$ and $2^{nd}$ days of three consecutive months in summer (June, July, and August) and predict for the $2^{\text{nd}}$ and $3^{rd}$ days of each month. We end up with around $500$k pixels (samples) per day. The output from the deep learning models (LSTM, ConvLSTM, Ours) is a probability $Y\in[0, 1]$. In addition, we visualize the predicted maps produced by FWI with the provided spatial resolution $8\text{ km} \times 8\text{ km}$. The output of FWI is clipped to the range $[0, 50]$ \cite{Prapas2}. The ignition points of large wildfires at those days are represented as black circles on the map. The first observation is that regardless of the coarse resolution of FWI, the predicted maps produced by the deep learning models are more reliable. While FWI relies on meteorological observations and models functional relationships, the results show that the modeled functional relationships are insufficient and do not reflect the complexity of the problem of forecasting wildfire. We also find that our proposed model with TE discards spots that result in a high false positive error rate while it keeps extreme ones (cf.\ the results for July in $2020$ and $2021$). Another important observation is that the LSTM model, which does not account for the spatial context, tends to produce heterogeneous predictions where neighboring pixels often have very different wildfire danger probabilities. Consequently, it generates many false positives. In contrast, our proposed model and ConvLSTM produce more homogeneous and clustered predictions. The respective performance metrics for Figs.~\ref{fig:6} and \ref{fig:7} are provided in Table \ref{tab:table5}.

\begin{table*}[t]
  \centering
  \caption{Recall (\%) metric per class for Figures \ref{fig:5} and \ref{fig:6}. (P) denotes the positive class and (N) the negative one. The number of positive samples is shown below the date. This is equivalent to the area (${\text{km}}^2$) that was burned by large wildfires at that day.\label{tab:table5}}
  \begin{tabular}{l *{12}c}
  \toprule
  & \multicolumn{2}{c}{02/06/2020} & \multicolumn{2}{c}{03/06/2020} & \multicolumn{2}{c}{02/07/2020} & \multicolumn{2}{c}{03/07/2020} & \multicolumn{2}{c}{02/08/2020} & \multicolumn{2}{c}{03/08/2020} \\  
 \# Positive samples & \multicolumn{2}{c}{0} & \multicolumn{2}{c}{0} & \multicolumn{2}{c}{2} & \multicolumn{2}{c}{0} & \multicolumn{2}{c}{71} & \multicolumn{2}{c}{25} \\
 \midrule
  Algorithm & P & N & P & N & P & N & P & N & P & N & P & N \\
  \midrule
   LSTM \cite{Prapas2} & - & 99.98 & - & 99.96 & 00.00 & 81.41 & - & 82.37 & 87.32 & 48.93 & \textbf{88.00} & 56.23 \\
  ConvLSTM \cite{Prapas2} & - & \textbf{100.00} & - & \textbf{100.00} & 00.00 & 89.72 & - & 92.46 & \textbf{92.96} & 65.09 & 80.00 & 72.52 \\
   2D/3D CNN & - & \textbf{100.00} & - & 99.99 & \textbf{100.00} & 89.51 & - & 90.33 & 73.24 & \textbf{66.74} & 80.00 & \textbf{76.17} \\
   2D/3D CNN w/ TE & - & \textbf{100.00} & - & \textbf{100.00} & 00.00 & \textbf{96.81} & - & \textbf{97.92} & 78.87 & 65.68 & 80.00 & 74.32 \\
   \midrule
 & \multicolumn{2}{c}{02/06/2021} & \multicolumn{2}{c}{03/06/2021} & \multicolumn{2}{c}{02/07/2021} & \multicolumn{2}{c}{03/07/2021} & \multicolumn{2}{c}{02/08/2021} & \multicolumn{2}{c}{03/08/2021} \\
 \# Positive samples & \multicolumn{2}{c}{0} & \multicolumn{2}{c}{0} & \multicolumn{2}{c}{36} & \multicolumn{2}{c}{22} & \multicolumn{2}{c}{679} & \multicolumn{2}{c}{1417} \\
 \midrule
  Algorithm & P & N & P & N & P & N & P & N & P & N & P & N \\
  \midrule
   LSTM \cite{Prapas2} & - & 98.23 & - & 96.44 & \textbf{100.00} & 70.58 & 68.18 & 83.14 & 82.03 & 32.54 & 93.30 & 23.88 \\
  ConvLSTM \cite{Prapas2} & - & 99.11 & - & 98.25 & \textbf{100.00} & 75.53 & \textbf{100.00} & 87.42 & 71.87 & 35.66 & 87.01 & 28.28 \\
   2D/3D CNN & -  & 99.38 & - & 98.08 & \textbf{100.00} & 84.36 & \textbf{100.00} & 88.97 & 83.80 & \textbf{39.17} & 99.29 & \textbf{33.75} \\
   2D/3D CNN w/ TE & - & \textbf{99.91} & - & \textbf{98.95} & \textbf{100.00} & \textbf{94.77} & \textbf{100.00} & \textbf{98.61} & \textbf{85.13} & 36.67 & \textbf{99.36} & 31.67 \\
   \bottomrule
   \end{tabular}
\end{table*}
\section{Ablation Study}
\label{sec:VI}

We finally evaluate two additional aspects. In Section \ref{sec:VI-B}, we analyze at which blocks of the proposed 2D/3D CNN LOAN is best added and the impact of the absolute temporal encoding with respect to the number of negative samples in Section~\ref{sec:VI-C}.         

\subsection{LOAN Position in the Model}\label{sec:VI-B}
\begin{table}[t]
\begin{center}
\caption{Ablation study of different position choices for the LOAN layer. All classification metrics are given in percent (\%). \label{tab:table6}}
\setlength\tabcolsep{3pt}
\begin{tabular}{l *{7}c}
\toprule
\multicolumn{7}{c}{Year $2019$ (val)}\\
\midrule
\multicolumn{3}{c}{Block} & \multirow{2.5}*{Precision($\uparrow$)} & \multirow{2.5}*{Recall($\uparrow$)} & \multirow{2.5}*{F1-score($\uparrow$)} & \multirow{2.5}*{AUROC($\uparrow$)} \\
\cmidrule{1-3}
$1^{\text{st}}$ & $2^{\text{nd}}$ & $3^{\text{rd}}$ \\
\midrule
$\times$ & $\times$ & $\times$ & 75.68 & 62.23 & 68.30 & 93.38 \\
\midrule
$\surd$ & $\times$ & $\times$ & 76.90 & \textbf{75.54} & 76.21 & 94.40 \\
\midrule
$\surd$ & $\surd$ & $\times$ & \textbf{79.64} & 74.62 & \textbf{77.05} & \textbf{94.52} \\
\midrule
$\surd$ & $\surd$ & $\surd$ & 73.92 & 70.85 & 72.35 & 93.68 \\
\midrule
\midrule
\multicolumn{7}{c}{Year $2020$-$2021$ (test)}\\
\midrule
\multicolumn{3}{c}{Block} & \multirow{2.5}*{Precision($\uparrow$)} & \multirow{2.5}*{Recall($\uparrow$)} & \multirow{2.5}*{F1-score($\uparrow$)} & \multirow{2.5}*{AUROC($\uparrow$)} \\
\cmidrule{1-3}
$1^{\text{st}}$ & $2^{\text{nd}}$ & $3^{\text{rd}}$ \\
\midrule
$\times$ & $\times$ & $\times$ & 83.35 & 79.98 & 81.63 & 97.11 \\
\midrule
$\surd$ & $\times$ & $\times$ & 80.90 & 78.72 & 79.80 & 97.22 \\
\midrule
$\surd$ & $\surd$ & $\times$ & \textbf{83.71} & \textbf{83.60} & \textbf{83.65} & \textbf{97.41} \\
\midrule
$\surd$ & $\surd$ & $\surd$ & 79.77 & 81.30 & 80.52 & 96.92 \\
\bottomrule
\end{tabular}
\end{center}
\end{table}

As shown in Fig.~\ref{fig:1}, the proposed network has three blocks and we add LOAN to the first and second block. We evaluate in Table \ref{tab:table6} different configurations where we add LOAN only to the first or to all three blocks. The results are reported without TE. If we add LOAN only to the first block, the performance increases for the year $2019$ but not for the years $2020$-$2021$ compared to our model without LOAN (first row). When adding LOAN to the first two blocks, we observe a consistent improvement for all years. For the year $2019$, Precision, Recall, F1-score, and AUROC are improved by $+3.96\%$, $+12.39\%$, $+8.75\%$, and $+1.14\%$, respectively, and by $+0.36\%$, $+3.62\%$, $+2.02\%$, and $+0.30\%$ for the years $2020$-$2021$, respectively. Adding LOAN to all three blocks performs worse than adding LOAN only to the first two blocks. This is due to the decrease of spatial resolution after each block by the pooling layers. In the third block the spatial resolution is too coarse for a location-specific modulation of the dynamic features.

\subsection{Absolute Temporal Encoding} \label{sec:VI-C}

\begin{figure}[h]
    \centering
    \includegraphics[width=3.4in]{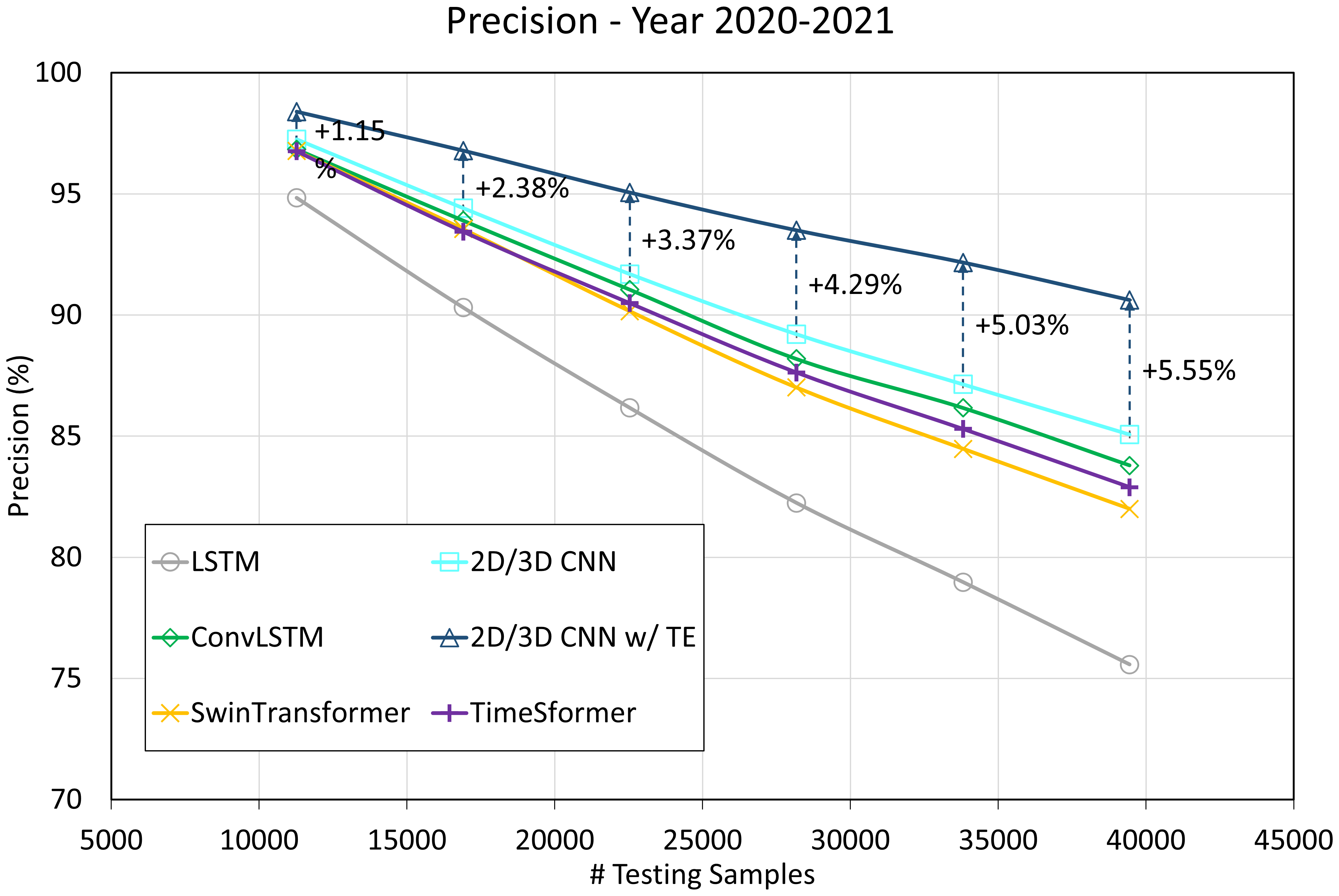}
    \caption{Impact of the number of testing samples on the precision. (TE) denotes the absolute temporal encoding.}
    \label{fig:9}
\end{figure}

\begin{figure}[!h]
    \centering
    \includegraphics[width=3.5in]{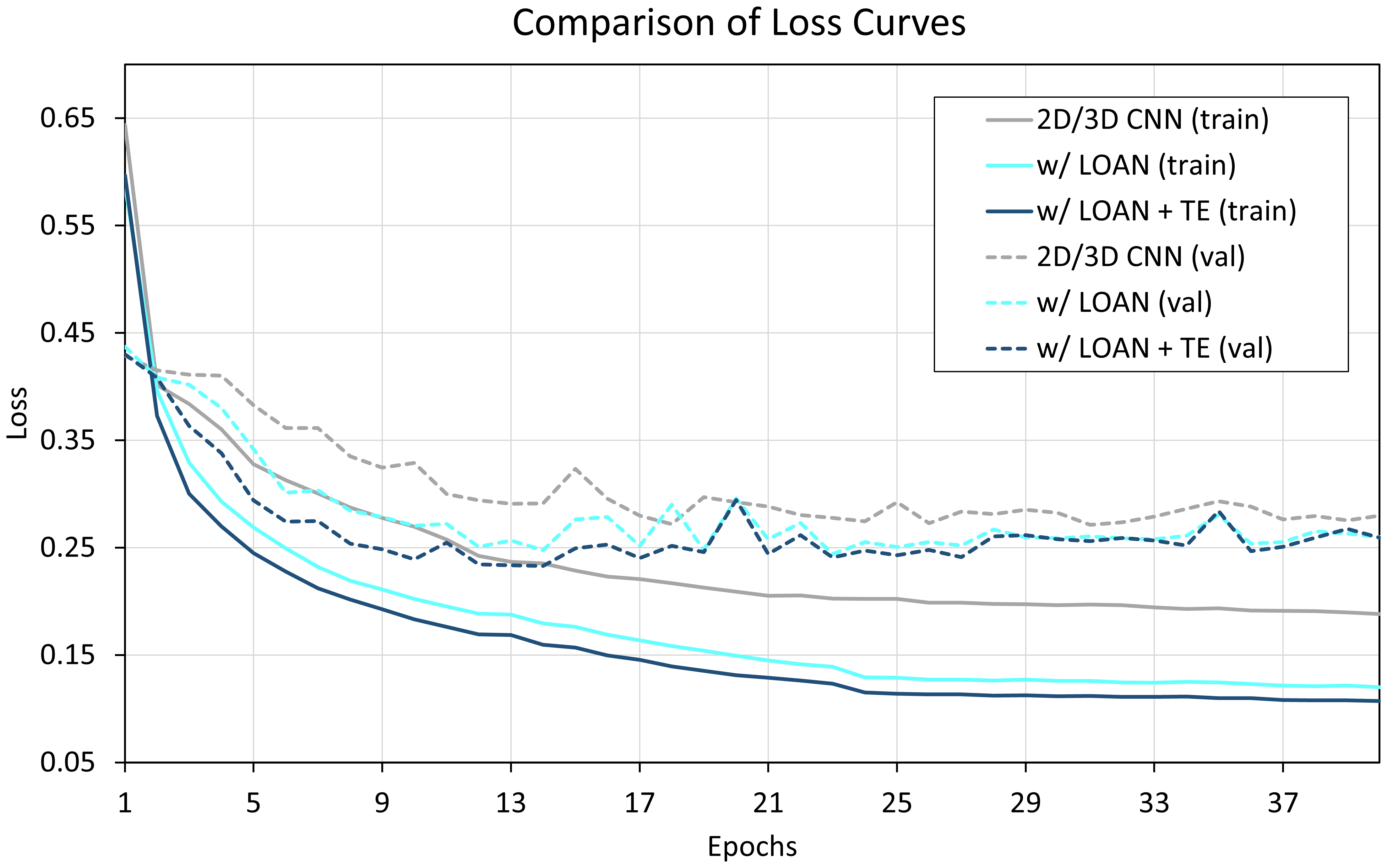}
    \caption{BCE loss during training for the training (solid) and validation (dashed) set. (LOAN) denotes the Location-aware Adaptive Normalization layer and (TE) denotes the absolute temporal encoding.}
    \label{fig:13}
\end{figure}

As we have seen in Tables \ref{tab:table1} and \ref{tab:table2}, absolute temporal encoding (TE) increases precision at the cost of lower recall. Depending on the use of the wildfire forecasting, recall or precision are more important. The precision also depends on the amount of negative samples. In order to show that TE gives consistently a higher precision, we varied the number of negative samples. In this experiment, we test on all positive samples in the test set (years $2020$-$2021$) and gradually increase the number of negative ones. As shown in Fig.~\ref{fig:9}, we start with a setup where the number of negative samples is equal to the number of positive samples, i.e., 5635 positive and 5635 negative samples. We then increase the number of negative samples. Since the number of negative samples increases, the precision decreases for all methods. Note that the recall does not change since the number of positive samples remains the same. As already observed in Table \ref{tab:table2}, LSTM has a very low precision. 2D/3D CNN with LOAN has in all settings a higher precision than ConvLSTM and a much higher recall as shown in Table \ref{tab:table2}. Transformer models on the other hand have less precision than ConvLSTM but provide an overall better recall as shown in Table \ref{tab:table2}. While adding TE decreases recall, it improves the precision substantially and the improvement increases when the number of negative samples increases. While other metrics like F1-score or AUROC combine precision and recall in a single measure, depending on the application a higher recall or a higher precision might be more important. If precision is more important, TE is very useful. If recall is more important, TE should not be used. We also point out that TE encodes only the day of the year since the dataset has a relatively small spatial extension ($10.2^{\circ} \text{Lon} \times 8^{\circ} \text{Lat}$). In case of larger datasets at continental scale, a consideration of the spatial location for the encoding would also become relevant as biogeographical regions occur \cite{NYBORG2022301}, which are characterized by different climate variabilities and anthropogenic drivers over time. Finally, we plot in Fig.~\ref{fig:13} the loss \eqref{eq:13} curve during training.
\section{Conclusion}
\label{sec:VII}
In this work, we proposed a new deep learning approach for wildfire danger forecasting. In contrast to previous works, we handle spatial (static) and spatio-temporal (dynamic) variables differently. Our model processes the spatial and spatio-temporal variables in two separated 2D/3D CNN branches to learn static and dynamic feature vectors. Moreover, we have introduced the Location-aware Adaptive Normalization layer, which modulates the activation maps in the dynamic branch conditionally on their respective static features to address the causal effect of static features on dynamic features. We furthermore integrated an absolute time encoding into the model. By encoding the calendar time, we make the model explicitly aware of the forecasting day. While the time encoding reduces the recall, it substantially increases the precision. We conducted our experiments on the FireCube dataset and demonstrated the effectiveness of our approach compared to several baselines in terms of Precision, F1-score, AUROC, and OA. Although our approach demonstrated a substantial improvement compared to previous works for wildfire forecasting, it still has some limitations. Despite the fact that our framework includes domain knowledge through the normalization layer and absolute time encoding, it does not incorporate physical knowledge about the Earth system. Furthermore, the FireCube dataset covers only parts of Eastern Mediterranean and the years 2009-2021. There is a need for more standardized datasets for wildfire forecasting at a continental scale and longer time periods. Finally, there may be hidden events that are correlated with climate variability and extreme weather conditions. It is an open question how these impact the forecast quality and if additional input variables will be needed to improve the forecast accuracy.

We believe that the proposed approach of dealing with spatial and spatio-temporal variables is also highly relevant for other remote sensing applications. 

\section*{Acknowledgments}
We would like to thank Ioannis Prapas and Spyros Kondylatos for providing the FireCube dataset.


\ifCLASSOPTIONcaptionsoff
  \newpage
\fi

\bibliographystyle{IEEEtran}
\bibliography{Ref}

\begin{IEEEbiography}[{\includegraphics[width=1in,height=1.25in,clip,keepaspectratio]{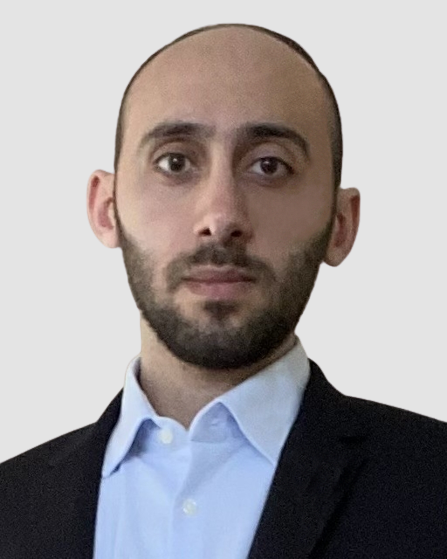}}]{Mohamad Hakam Shams Eddin} is a PhD student at the University of Bonn, Germany. He received his Dipl.-Ing. degree in topographic engineering from the University of Aleppo, Syria, in 2015 and his M.Sc. degree in geomatics engineering from the University of Stuttgart, Germany, in 2019. His research interests include deep learning, remote sensing, and anomaly detection.
\end{IEEEbiography}

\begin{IEEEbiography}[{\includegraphics[width=1in,height=1.25in,clip,keepaspectratio]{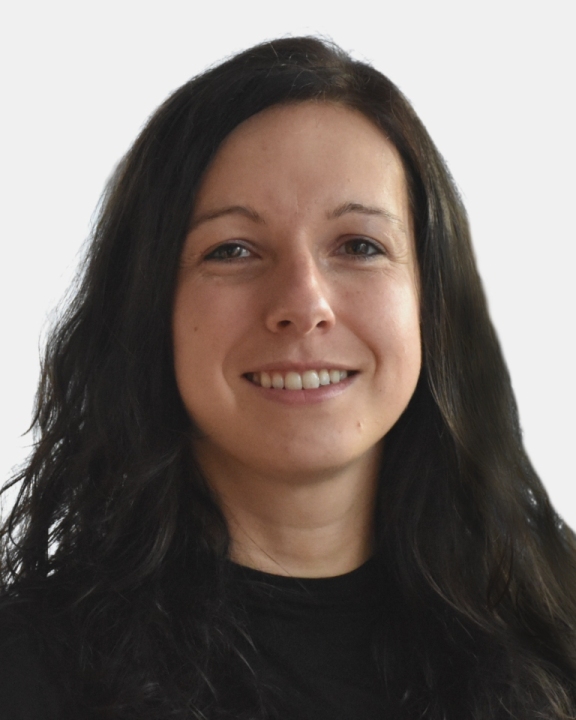}}]{Ribana Roscher}
(Member, IEEE) received the Dipl.Ing. and Ph.D. degrees in geodesy from the University of Bonn, Bonn, Germany, in 2008 and 2012, respectively. Until 2022, she was a Junior Professor of Remote Sensing with the University of Bonn. Before she was a Postdoctoral Researcher with the University of Bonn, the Julius-Kuehn Institute, Siebeldingen, Germany, Freie Universitaet Berlin, Berlin, Germany, and the Humboldt Innovation, Berlin. In 2015, she was a Visiting
Researcher with the Fields Institute, Toronto, ON, Canada.
Since 2022, Ribana Roscher is a professor of Data Science for Crop Systems at the University of Bonn, Bonn, Germany. Currently, she leads the Data Science research area at IBG-2, Forschungszentrum Jülich.  \end{IEEEbiography}

\begin{IEEEbiography}[{\includegraphics[width=1in,height=1.25in,clip,keepaspectratio]{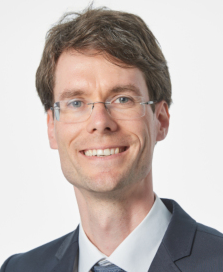}}]{Juergen Gall}
(Member, IEEE) received his B.Sc. and M.Sc. degrees in mathematics from the University of Wales Swansea, UK and from the University of Mannheim, Germany, in 2004 and 2005, respectively. In 2009, he obtained a Ph.D. in computer science from the Saarland University and the Max Planck Institut fuer Informatik. He was a postdoctoral researcher at the Computer Vision Laboratory, ETH Zurich, Switzerland from 2009 until 2012 and a senior research scientist at the Max Planck Institute for Intelligent Systems in Tuebingen from 2012 until 2013. Since 2013, Juergen Gall is a professor at the University of Bonn and head of Computer Vision Group. He is furthermore member of the Lamarr Institute for Machine Learning and Artificial Intelligence.
\end{IEEEbiography}

\vfill

\end{document}